\documentclass[letterpaper]{article} 
\usepackage[preprint]{aaai2027}  
\usepackage[hyphens]{url}  
\usepackage{graphicx} 
\usepackage{natbib}  
\usepackage{caption} 
\usepackage{algorithm}
\usepackage{algorithmic}
\usepackage{amsmath}
\usepackage{amssymb}
\usepackage{diagbox}
\usepackage[most]{tcolorbox}
\usepackage{needspace}
\usepackage{array}

\usepackage{newfloat}
\usepackage{listings}
\DeclareCaptionStyle{ruled}{labelfont=normalfont,labelsep=colon,strut=off} 
\floatstyle{ruled}
\newfloat{listing}{tb}{lst}{}
\floatname{listing}{Listing}

\usepackage{xcolor}
\newcolumntype{M}[1]{>{\raggedright\arraybackslash}m{#1}}
\newcolumntype{C}[1]{>{\centering\arraybackslash}m{#1}}
\newfloat{promptlisting}{tb}{lop}{}
\floatname{promptlisting}{Prompt}

\definecolor{promptframe}{HTML}{3B4A5A}
\definecolor{prompttitle}{HTML}{25313D}
\definecolor{promptback}{HTML}{F7F8FA}
\definecolor{promptnote}{HTML}{EEF2F6}
\definecolor{promptheadingcolor}{HTML}{7A2632}
\definecolor{promptmetaframe}{HTML}{4D6F93}
\definecolor{promptbodyframe}{HTML}{9A6A2F}
\definecolor{promptmetaback}{HTML}{F2F6FB}
\definecolor{promptbodyback}{HTML}{FFF8EA}
\definecolor{promptsubheadingcolor}{HTML}{34495E}

\newcommand{\promptcaption}[2][]{%
    \par\smallskip
    \noindent\begin{minipage}{\linewidth}
    \captionsetup{skip=2pt}%
    \captionof{promptlisting}{#2}%
    \if\relax\detokenize{#1}\relax\else\label{#1}\fi
    \end{minipage}
    \par
}

\newcommand{\promptheading}[1]{%
    \par\smallskip
    \noindent{\color{promptsubheadingcolor}\bfseries #1}\par
}

\newcommand{\promptmainheading}[1]{%
    \par\smallskip
    \noindent{\color{promptheadingcolor}\large\bfseries #1}\par
}

\newenvironment{promptcontent}{%
    \begingroup
    \parindent=1.1em
}{%
    \par
    \endgroup
}

\newtcolorbox{promptnotes}[1][]{
    enhanced,
    breakable,
    bicolor,
    sharp corners,
    boxrule=0.45pt,
    colback=promptnote,
    colframe=promptframe,
    left=5pt,
    right=5pt,
    top=4pt,
    bottom=4pt,
    fontupper=\footnotesize,
    #1
}

\newcounter{operatorpromptcounter}

\newtcolorbox{operatorpromptbox}[1][]{
    enhanced,
    breakable,
    boxrule=0.55pt,
    sharp corners,
    colback=promptback,
    colframe=black,
    left=6pt,
    right=6pt,
    top=5pt,
    bottom=5pt,
    before skip=3pt,
    after skip=3pt,
    before upper={%
        \setcounter{operatorpromptcounter}{0}%
        \small
        \setlength{\parskip}{3pt}%
    },
    #1
}

\newcommand{\operatorprompt}[2]{%
    \refstepcounter{operatorpromptcounter}%
    \begin{tcolorbox}[
        enhanced,
        breakable,
        boxrule=0.5pt,
        arc=1.2mm,
        outer arc=1.2mm,
        colback=promptback,
        colframe=promptheadingcolor,
        left=6pt,
        right=6pt,
        top=4pt,
        bottom=4pt,
        before skip=3pt,
        after skip=3pt
    ]
    \noindent{\color{promptheadingcolor}\bfseries Operator~\theoperatorpromptcounter: #1}\par
    \begin{promptcontent}%
    #2%
    \end{promptcontent}%
    \end{tcolorbox}%
}

\newenvironment{promptmetablock}[1][]{%
    \begingroup
    \setlength{\leftskip}{12pt}%
    \setlength{\rightskip}{12pt}%
    \setlength{\parindent}{0pt}%
    \setlength{\parskip}{2pt}%
    \vspace{3pt}%
}{%
    \par
    \vspace{3pt}%
    \endgroup
}

\newenvironment{promptinput}[1][]{%
    \promptmainheading{Input}%
}{%
}

\newenvironment{promptoutput}[1][]{%
    \promptmainheading{Output}%
}{%
}

\newenvironment{prompttext}[1][]{%
    \begingroup
    \setlength{\leftskip}{12pt}%
    \setlength{\rightskip}{12pt}%
    \promptmainheading{Prompt Text}%
}{%
    \par
    \endgroup
}

\makeatletter
\newcommand{\promptdrawmetacomplete}{%
    \path[fill=promptmetaback,draw=promptmetaframe,line width=0.55pt,rounded corners=1.5mm]
        ([xshift=10pt,yshift=-9pt]interior.north west) rectangle
        ([xshift=-10pt,yshift=3pt]segmentation.east);
}

\newcommand{\promptdrawmetafirst}{%
    \path[fill=promptmetaback,rounded corners=1.5mm]
        ([xshift=10pt,yshift=-9pt]interior.north west) rectangle
        ([xshift=-10pt]interior.south east);
    \draw[promptmetaframe,line width=0.55pt,rounded corners=1.5mm]
        ([xshift=10pt]interior.south west) --
        ([xshift=10pt,yshift=-9pt]interior.north west) --
        ([xshift=-10pt,yshift=-9pt]interior.north east) --
        ([xshift=-10pt]interior.south east);
}

\newcommand{\promptdrawmetamiddle}{%
    \path[fill=promptmetaback]
        ([xshift=10pt]interior.north west) rectangle
        ([xshift=-10pt]interior.south east);
    \draw[promptmetaframe,line width=0.55pt]
        ([xshift=10pt]interior.north west) --
        ([xshift=10pt]interior.south west);
    \draw[promptmetaframe,line width=0.55pt]
        ([xshift=-10pt]interior.north east) --
        ([xshift=-10pt]interior.south east);
}

\newcommand{\promptdrawmetalast}{%
    \path[fill=promptmetaback]
        ([xshift=10pt]interior.north west) rectangle
        ([xshift=-10pt,yshift=5pt]interior.south east);
    \draw[promptmetaframe,line width=0.55pt,rounded corners=1.5mm]
        ([xshift=10pt]interior.north west) --
        ([xshift=10pt,yshift=5pt]interior.south west) --
        ([xshift=-10pt,yshift=5pt]interior.south east) --
        ([xshift=-10pt]interior.north east);
}

\newcommand{\promptdrawmetalasttosegmentation}{%
    \path[fill=promptmetaback]
        ([xshift=10pt]interior.north west) rectangle
        ([xshift=-10pt,yshift=3pt]segmentation.east);
    \draw[promptmetaframe,line width=0.55pt,rounded corners=1.5mm]
        ([xshift=10pt]interior.north west) --
        ([xshift=10pt,yshift=3pt]segmentation.west) --
        ([xshift=-10pt,yshift=3pt]segmentation.east) --
        ([xshift=-10pt]interior.north east);
}

\newcommand{\promptdrawbodycomplete}{%
    \path[fill=promptbodyback,draw=promptbodyframe,line width=0.55pt,rounded corners=1.5mm]
        ([xshift=10pt,yshift=-3pt]segmentation.west) rectangle
        ([xshift=-10pt,yshift=5pt]interior.south east);
}

\newcommand{\promptdrawbodyfirst}{%
    \path[fill=promptbodyback,rounded corners=1.5mm]
        ([xshift=10pt,yshift=-3pt]segmentation.west) rectangle
        ([xshift=-10pt]interior.south east);
    \draw[promptbodyframe,line width=0.55pt,rounded corners=1.5mm]
        ([xshift=10pt]interior.south west) --
        ([xshift=10pt,yshift=-3pt]segmentation.west) --
        ([xshift=-10pt,yshift=-3pt]segmentation.east) --
        ([xshift=-10pt]interior.south east);
}

\newcommand{\promptdrawbodymiddle}{%
    \path[fill=promptbodyback]
        ([xshift=10pt]interior.north west) rectangle
        ([xshift=-10pt]interior.south east);
    \draw[promptbodyframe,line width=0.55pt]
        ([xshift=10pt]interior.north west) --
        ([xshift=10pt]interior.south west);
    \draw[promptbodyframe,line width=0.55pt]
        ([xshift=-10pt]interior.north east) --
        ([xshift=-10pt]interior.south east);
}

\newcommand{\promptdrawbodylast}{%
    \path[fill=promptbodyback]
        ([xshift=10pt]interior.north west) rectangle
        ([xshift=-10pt,yshift=5pt]interior.south east);
    \draw[promptbodyframe,line width=0.55pt,rounded corners=1.5mm]
        ([xshift=10pt]interior.north west) --
        ([xshift=10pt,yshift=5pt]interior.south west) --
        ([xshift=-10pt,yshift=5pt]interior.south east) --
        ([xshift=-10pt]interior.north east);
}

\newcommand{\promptdrawbodyallfromsegmentation}{%
    \path[fill=promptbodyback,draw=promptbodyframe,line width=0.55pt,rounded corners=1.5mm]
        ([xshift=10pt,yshift=-3pt]segmentation.west) rectangle
        ([xshift=-10pt,yshift=5pt]interior.south east);
}

\newcommand{\promptifsegmentation}[2]{%
    \ifcsname pgf@sh@ns@segmentation\endcsname
        #1%
    \else
        #2%
    \fi
}

\newcommand{\promptunderlayfirst}{%
    \promptifsegmentation{%
        \ifdim\tcb@h@upper>0pt\relax
            \promptdrawmetacomplete
        \fi
        \ifdim\tcb@h@lower>0pt\relax
            \promptdrawbodyfirst
        \fi
    }{%
        \ifdim\tcb@h@upper>0pt\relax
            \promptdrawmetafirst
        \else
            \ifdim\tcb@h@lower>0pt\relax
                \promptdrawbodymiddle
            \fi
        \fi
    }%
}

\newcommand{\promptunderlaymiddle}{%
    \promptifsegmentation{%
        \ifdim\tcb@h@upper>0pt\relax
            \ifdim\tcb@h@lower>0pt\relax
                \promptdrawmetalasttosegmentation
                \promptdrawbodyfirst
            \else
                \promptdrawmetamiddle
            \fi
        \else
            \ifdim\tcb@h@lower>0pt\relax
                \promptdrawbodymiddle
            \fi
        \fi
    }{%
        \ifdim\tcb@h@upper>0pt\relax
            \promptdrawmetamiddle
        \else
            \ifdim\tcb@h@lower>0pt\relax
                \promptdrawbodymiddle
            \fi
        \fi
    }%
}

\newcommand{\promptunderlaylast}{%
    \promptifsegmentation{%
        \ifdim\tcb@h@upper>0pt\relax
            \ifdim\tcb@h@lower>0pt\relax
                \promptdrawmetalasttosegmentation
                \promptdrawbodyallfromsegmentation
            \else
                \promptdrawmetalast
            \fi
        \else
            \ifdim\tcb@h@lower>0pt\relax
                \promptdrawbodylast
            \fi
        \fi
    }{%
        \ifdim\tcb@h@upper>0pt\relax
            \promptdrawmetalast
        \else
            \ifdim\tcb@h@lower>0pt\relax
                \promptdrawbodylast
            \fi
        \fi
    }%
}
\makeatother

\newtcolorbox[auto counter]{promptcard}[2][]{
    enhanced,
    breakable,
    bicolor,
    sharp corners,
    boxrule=0.75pt,
    colback=promptback,
    colbacklower=promptback,
    colframe=promptframe,
    coltitle=white,
    colbacktitle=prompttitle,
    fonttitle=\bfseries\footnotesize,
    title={Prompt~\thetcbcounter: #2},
    notitle after break,
    pad at break=2pt,
    left=6pt,
    right=6pt,
    top=5pt,
    bottom=5pt,
    segmentation style={draw=none},
    underlay unbroken={%
        \promptdrawmetacomplete
        \promptdrawbodycomplete
    },
    underlay first={%
        \promptunderlayfirst
    },
    underlay middle={%
        \promptunderlaymiddle
    },
    underlay last={%
        \promptunderlaylast
    },
    before skip=8pt,
    fontupper=\small,
    before upper={\setlength{\parindent}{0pt}\setlength{\parskip}{2pt}},
    #1
}

\newtcolorbox[auto counter]{requirementcard}[2][]{
    enhanced,
    breakable,
    sharp corners,
    boxrule=0.75pt,
    colback=promptback,
    colframe=promptframe,
    coltitle=white,
    colbacktitle=promptheadingcolor,
    fonttitle=\bfseries\footnotesize,
    title={Requirement~\thetcbcounter: #2},
    notitle after break,
    pad at break=2pt,
    left=7pt,
    right=7pt,
    top=5pt,
    bottom=5pt,
    before skip=8pt,
    after skip=8pt,
    fontupper=\small,
    before upper={\setlength{\parindent}{0pt}\setlength{\parskip}{3pt}},
    #1
}

\newcommand{\reqheading}[1]{%
    \par\smallskip
    \noindent{\color{promptheadingcolor}\bfseries #1}\par
}

\newcommand{\reqitem}[3]{%
    \par\noindent\textbf{#1.} \textit{#2}: #3\par
}

\definecolor{codeback}{HTML}{F8FAFC}
\definecolor{codeframe}{HTML}{4B5D6E}
\definecolor{codekeyword}{HTML}{1F5E99}
\definecolor{codestring}{HTML}{7A2632}
\definecolor{codecomment}{HTML}{6B7280}
\definecolor{codenumber}{HTML}{7C8794}

\lstdefinestyle{formulationpython}{
    language=Python,
    basicstyle=\scriptsize\ttfamily,
    keywordstyle=\color{codekeyword}\bfseries,
    stringstyle=\color{codestring},
    commentstyle=\color{codecomment}\itshape,
    numbers=left,
    numberstyle=\tiny\color{codenumber},
    numbersep=5pt,
    showstringspaces=false,
    breaklines=true,
    breakatwhitespace=false,
    tabsize=4,
    columns=fullflexible,
    keepspaces=true,
    frame=none,
    aboveskip=0pt,
    belowskip=7pt
}

\newcounter{formulationcounter}
\newcommand{\formulationsubheading}[1]{%
    \par\medskip\noindent{\bfseries #1}\par\nobreak\vspace{2pt}%
}
\newcommand{\formulationtitle}[2][]{%
    \refstepcounter{formulationcounter}%
    \if\relax\detokenize{#1}\relax\else\label{#1}\fi
    \par\vspace{5pt}%
    \par\noindent\hspace*{-4.5pt}\makebox[0pt][l]{\colorbox{prompttitle}{%
        \makebox[\dimexpr\linewidth+9pt-2\fboxsep\relax][l]{%
            \color{white}\bfseries\footnotesize\rule[-4pt]{0pt}{10pt}Formulation~\theformulationcounter: #2%
        }%
    }}%
    \par\nobreak\vspace{-0.45pt}%
}
\lstnewenvironment{formulationcode}[1][]{%
    \lstset{
        style=formulationpython,
        backgroundcolor=\color{codeback},
        frame=single,
        framerule=0.45pt,
        rulecolor=\color{codeframe},
        framesep=4pt,
        xleftmargin=0pt,
        xrightmargin=0pt,
        #1
    }%
}{}

\usepackage{booktabs}

\title{Search Hardness-Aware LLM-Based Problem Formulation for Expensive Simulation-Driven Design}
\author{
    Yuchen Li\textsuperscript{\rm 1},
    Handing Wang\textsuperscript{\rm 1}\corresponding,
    Bing Xue\textsuperscript{\rm 2},
    Mengjie Zhang\textsuperscript{\rm 2}
}
\affiliations{
    \textsuperscript{\rm 1}School of Artificial Intelligence, Xidian University, China\\
    \textsuperscript{\rm 2}School of Engineering and Computer Science, Victoria University of Wellington, New Zealand\\
    ycli\_7@stu.xidian.edu.cn, hdwang@xidian.edu.cn\\
    bing.xue@ecs.vuw.ac.nz, mengjie.zhang@ecs.vuw.ac.nz
}

\begin{document}

\maketitle

\begin{abstract}
Expensive simulation-driven design is widely used in engineering to identify requirement-satisfying designs with as few high-fidelity simulations as possible.
Most existing efforts address this challenge by improving optimization algorithms under fixed formulations, yet the formulation itself shapes the search landscape by defining the objectives and constraints optimized by the solver.
Recent LLM-based automatic problem formulation methods generate formulations from natural-language requirements, but they mainly focus on design-intent alignment and overlook whether the formulation induces an efficient search process.
To address this limitation, we propose SHA-PF, a search hardness-aware LLM-based problem formulation framework.
We find that a formulation is more likely to guide efficient search when it prioritizes rare samples with greater progress potential.
Based on this finding, SHA-PF defines a formulation search objective guided by search hardness, scoring each candidate formulation according to the priority.
SHA-PF then searches the formulation space under this objective through LLM-based generation, repair, and evolutionary refinement.
Experiments on the real-world multi-objective benchmark and five expensive antenna design benchmarks show that the formulations discovered by SHA-PF require significantly fewer evaluations to reach the design requirements than other baselines.
\end{abstract}

%
\begin{figure}[t]
    \centering
    \includegraphics[width=0.99\linewidth]{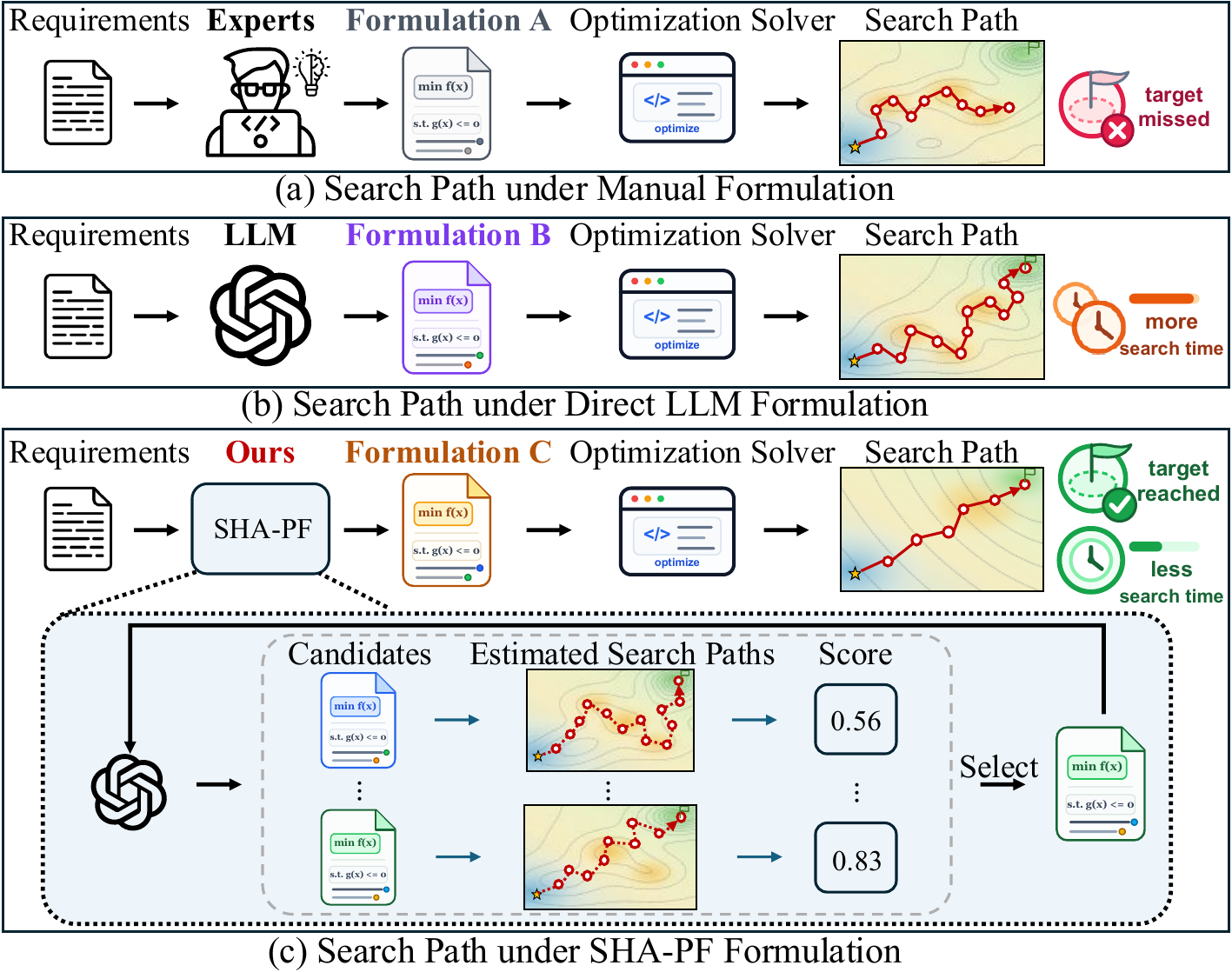} 

    \caption{Comparison of search paths under different problem formulation methods.}
    \label{fig1:search path compare}
\end{figure}
\section{Introduction}

Expensive simulation-driven design has been widely used to improve the performance of complex engineering systems, such as airfoil design~\cite{wissink2025multi,liMachineLearningAerodynamic2022}, and microelectronic design~\cite{NEURIPS2024_fb23cf87}.
In such problems, a time-consuming high-fidelity simulator is usually required to evaluate the performance response of a candidate design parameter.
The goal is to find design parameters whose responses satisfy natural-language requirements within a limited evaluation budget.
In practice, the design process usually involves two stages: problem formulation and optimization. 
The former converts natural-language requirements into a solvable optimization problem, while the latter searches for design parameters under the formulated optimization problem.

Most existing studies focus on developing surrogate-assisted optimization algorithms to reduce high-fidelity simulations due to the high cost of simulation evaluations~\cite{jinDataDrivenEvolutionaryOptimization2019}, where surrogate models predict promising \mbox{candidates} for expensive evaluations.
However, they typically rely on a strong assumption that the given problem formulation is correct and effective. 
The formulation defines the evaluation metrics of the solutions, such as objective functions and constraints, thereby shaping the search-space landscape and determining the search path from starting solutions to target solutions, as illustrated in Figure~\ref{fig1:search path compare}.
In practice, such formulations are usually manually designed by domain experts based on their knowledge and experience, making the process time-consuming and labor-intensive.

These limitations have motivated research on automated problem formulation~\cite{zhangSystematicSurveyLarge2025}.
Recent advances in large language models (LLMs), such as GPT~\cite{openaiGPT4TechnicalReport2024}, Gemini~\cite{teamGeminiFamilyHighly2025}, and DeepSeek~\cite{deepseek-aiDeepSeekV3TechnicalReport2025}, have further advanced this direction. 
Existing methods mainly focus on converting vague natural-language requirements into precise mathematical formulations, and can generally be divided into prompting-based and learning-based methods. Prompting-based methods, such as Chain-of-Experts~\cite{xiaoChainofExpertsWhenLLMs2023} and OptiMUS~\cite{ahmaditeshniziOptiMUSScalableOptimization2024}, rely on carefully designed prompting strategies or multi-agent collaboration and can be quickly deployed without additional training. 
Learning-based methods, such as LLaMoCo~\cite{maLLaMoCoInstructionTuning2024}, ORLM~\cite{huangORLMCustomizableFramework2025}, LLMOPT~\cite{jiangLLMOPTLearningDefine2024}, SIRL~\cite{chenSolverInformedRLGrounding2025}, and APF~\cite{li2026solverindependentautomatedproblemformulation}, usually improve reliability in complex scenarios through data synthesis and fine-tuning. 
Although these methods improve the formal accuracy of converting natural-language requirements into mathematical formulations, they mainly ensure that the shaped search-space landscape defines the correct target region.

In expensive simulation-driven design, even when the target region is correctly defined, there may exist many search paths from the starting solutions to that region. 
Due to the different landscape hardness along these paths, they require different numbers of expensive simulation evaluations and directly affect search efficiency.
Given the direct impact of search efficiency on design cycle and cost, automatically discovering formulations that induce more efficient search paths becomes essential for expensive simulation-driven design.

To address this challenge, we propose SHA-PF, a search hardness-aware LLM-based problem formulation framework that uses only the initially evaluated designs and their simulation responses to discover formulations that enable higher search efficiency.
We find that the initially evaluated data contain early partial paths toward the target solution, where rarely observed paths indicate harder-to-reach and more informative search directions.
If a formulation assigns higher priority to these rarely observed but promising paths, the optimizer is more likely to continue exploring along them and reach the target solution with fewer expensive evaluations than it would by following commonly observed paths.
Based on this finding, SHA-PF constructs an anchor-state-based evaluation criterion to score candidate formulations and discover efficient problem formulations.

The main contributions of this paper are as follows:
(1) We reveal that rarely observed but promising search paths in the initial simulation data are more valuable for continued search than commonly observed paths.
(2) We propose an anchor-state-based formulation evaluation criterion that favors formulations assigning higher priority to rarely observed but promising paths, thereby encouraging the optimizer to continue exploring along these paths.
(3) We propose SHA-PF, which integrates the proposed evaluation criterion with LLM-based generation, validity repair, and iterative refinement to discover efficient formulations.
We validate the effectiveness of SHA-PF on a real-world multi-objective benchmark and antenna design tasks. 
The experimental results show that SHA-PF can significantly reduce the number of expensive simulation evaluations.

\section{Related Work}
\subsection{Expensive Simulation-Driven Optimization}

In simulation-driven optimization, evaluating each candidate design requires costly simulations, which makes the affordable number of evaluations very limited.
This bottleneck has motivated surrogate-assisted optimization methods, which use inexpensive surrogate models to reduce costly true evaluations~\cite{jinDataDrivenEvolutionaryOptimization2019,wangRecentAdvancesBayesian2023}.

Surrogate-assisted evolutionary algorithms (SAEAs) tackle expensive optimization from a population-based perspective, using surrogate models to guide evolutionary search and reduce exact evaluations~\cite{liuSurveyLearnableEvolutionary2023a}. 
Representative methods include GPEME for medium-scale single-objective expensive optimization, which uses Gaussian-process surrogates to prescreen promising offspring~\cite{liuGaussianProcessSurrogate2014}, and K-RVEA for expensive many-objective optimization, which couples Kriging models with reference-vector-guided selection~\cite{chughSurrogateAssistedReferenceVector2018}. 
Bayesian optimization (BO) is another major paradigm, using probabilistic surrogates and acquisition functions to decide which candidates to evaluate. 
TuRBO scales BO to high-dimensional black-box problems through local trust regions~\cite{erikssonScalableGlobalOptimization2019}. 
SCBO extends trust-region BO to constrained optimization by balancing improvement and feasibility~\cite{erikssonScalableConstrainedBayesian2021a}. 
qEHVI addresses parallel multi-objective BO with differentiable hypervolume-based acquisitions~\cite{daultonDifferentiableExpectedHypervolume2020}. 
qNEHVI extends this framework to noisy multi-objective evaluations~\cite{daultonParallelBayesianOptimization2021}.

Despite their differences, both SAEAs and BO usually assume that a well-defined optimization problem is already available before the search begins. 
However, problem formulation itself is an indispensable step in real simulation-driven design. 
For example, the unmanned underwater vehicle design task formalizes vehicle-dimension search as a constrained multi-objective optimization problem evaluated with low and high-fidelity computational fluid dynamics models~\cite{alamDesignOptimizationUnmanned2017}, while SGSO formalizes filtering patch antenna design as an expensive constrained optimization problem under electromagnetic simulations~\cite{xuSelfGuidedSimulationDrivenOptimization2026}. 
These methods rely on a sufficiently accurate and effective formulation, but they do not address how such a formulation should be automatically constructed from natural-language design requirements. 
This leaves automated problem formulation as a necessary problem for simulation-driven optimization.

\subsection{LLMs for Problem Formulation}
The NL4Opt benchmark and competition provided an early standardized testbed by decomposing natural-language optimization modeling into semantic-entity recognition and formulation generation tasks~\cite{ramamonjisonAugmentingOperationsResearch2022,ramamonjisonNL4OptCompetitionFormulating2023}. 
Subsequent prompt-based and agentic methods improve formulation by decomposing the modeling process, coordinating specialized agents, or supporting interactive refinement. 
Chain-of-Experts and OptiMUS use multiple LLM agents to coordinate modeling, programming, and verification~\cite{xiaoChainofExpertsWhenLLMs2023,ahmaditeshniziOptiMUSScalableOptimization2024}. 
OptLLM supports multi-round dialogues for gradually refining the modeling and solving process, while MeetMate studies interactive decision support in meeting scheduling by combining LLM-based preference construction with constraint-programming-based preference incorporation~\cite{zhangSolvingGeneralNaturalLanguageDescription2024,lawlessWantItThat2024}.

Recent work has further shifted from prompt engineering to learning-based formulation. 
ORLM synthesizes large-scale triples of natural-language problems, formal models, and solver code to fine-tune open-source LLMs~\cite{huangORLMCustomizableFramework2025}. 
OptMATH and ReSocratic improve synthetic data generation through bidirectional or inverse modeling pipelines~\cite{luOptMATHScalableBidirectional2025a,yangOptiBenchMeetsReSocratic2025}. 
LLMOPT trains LLMs with optimization-oriented instructions to improve both problem definition and solver-code generation across optimization problem types~\cite{jiangLLMOPTLearningDefine2024}, while SIRL introduces solver-verifiable reinforcement learning to improve formulation reliability~\cite{chenSolverInformedRLGrounding2025}. 
APF extends this line of work to high-cost simulation-driven design, where formulation learning must proceed without solver-based verification because each evaluation requires expensive simulation~\cite{li2026solverindependentautomatedproblemformulation}.
However, these studies mainly evaluate whether the generated formulation is correct, executable, or solver-verifiable, rather than whether it induces an efficient optimization process. 
This limitation is critical in expensive simulation-driven design, where different valid formulations may encode the same design intent but lead to search paths with substantially different simulation costs.

\section{Method}

\subsection{Search-Efficient Formulation}

In expensive simulation-driven design, each design candidate 
$\mathbf{x}\in\mathcal{X}$ is evaluated by a costly simulator 
$\mathcal{S}:\mathcal{X}\rightarrow\mathcal{Y}$, yielding a simulated response
$\mathbf{y}=\mathcal{S}(\mathbf{x})$.
Since each simulator call can be computationally expensive, the number of evaluations is limited by a budget $B$.

The design intent is specified by a natural-language requirement 
$\mathcal{R}=\{r_1,r_2,\ldots,r_K\}$, where each $r_k$ denotes a design criterion. 
For a response $\mathbf{y}\in\mathcal{Y}$, let 
$s_k(\mathbf{y})\in\{0,1\}$ indicate whether $\mathbf{y}$ satisfies criterion $r_k$, and define its satisfaction state as
\begin{equation}
    \mathbf{s}(\mathbf{y})
    =
    \left(
    s_1(\mathbf{y}),\ldots,s_K(\mathbf{y})
    \right)
    \in \{0,1\}^{K}.
\end{equation}
The requirement-satisfying response set is 
$\mathcal{Y}_{\mathcal{R}}^{+}
=
\{\mathbf{y}\in\mathcal{Y}\mid s_k(\mathbf{y})=1,\; \forall k=1,\ldots,K\}$.
The design goal is to find $\mathbf{x}\in\mathcal{X}$ such that 
$\mathcal{S}(\mathbf{x})\in\mathcal{Y}_{\mathcal{R}}^{+}$.

Since $\mathcal{R}$ is specified in natural language, it must be instantiated as an executable formulation $\mathcal{F}\in\Phi$, where $\Phi$ denotes the candidate formulation space. 
Let $\mathcal{X}_{\mathcal{F}}^{*}$ denote the ideal solution set induced by $\mathcal{F}$. 
The set of requirement-aligned formulations is defined as
\begin{equation}
    \Phi_{\mathcal{R}}
    =
    \left\{
    \mathcal{F}\in\Phi
    \mid
    \mathcal{S}(\mathbf{x})\in\mathcal{Y}_{\mathcal{R}}^{+},
    \; \forall \mathbf{x}\in\mathcal{X}_{\mathcal{F}}^{*}
    \right\}.
\end{equation}

Given a requirement-aligned formulation $\mathcal{F}\in\Phi_{\mathcal{R}}$, 
an optimizer $\mathcal{A}$ sequentially queries the simulator under budget $B$, 
forming a trajectory from lower-satisfaction states toward higher-satisfaction states:
\begin{equation}
    \mathcal{T}_{\mathcal{F}}^{\mathcal{A}}
    =
    \left\{
    (\mathbf{x}_{t},\mathbf{y}_{t})
    \right\}_{t=1}^{T},
    \quad
    \mathbf{y}_{t}=\mathcal{S}(\mathbf{x}_{t}).
\end{equation}
The trajectory ends when it first reaches $\mathcal{Y}_{\mathcal{R}}^{+}$ or when the budget is exhausted.
The search cost is defined as 
$\tau_{\mathcal{A}}(\mathcal{F})=T$ if 
$\mathbf{y}_{T}\in\mathcal{Y}_{\mathcal{R}}^{+}$, and 
$\tau_{\mathcal{A}}(\mathcal{F})=B+1$ otherwise.
Thus, we aim to find a formulation $\mathcal{F}\in\Phi_{\mathcal{R}}$ that minimizes $\tau_{\mathcal{A}}(\mathcal{F})$.

\subsection{Finding: Initial Data Reveal Search Hardness}
\begin{figure}[t]
    \centering
    \includegraphics[width=0.99\linewidth]{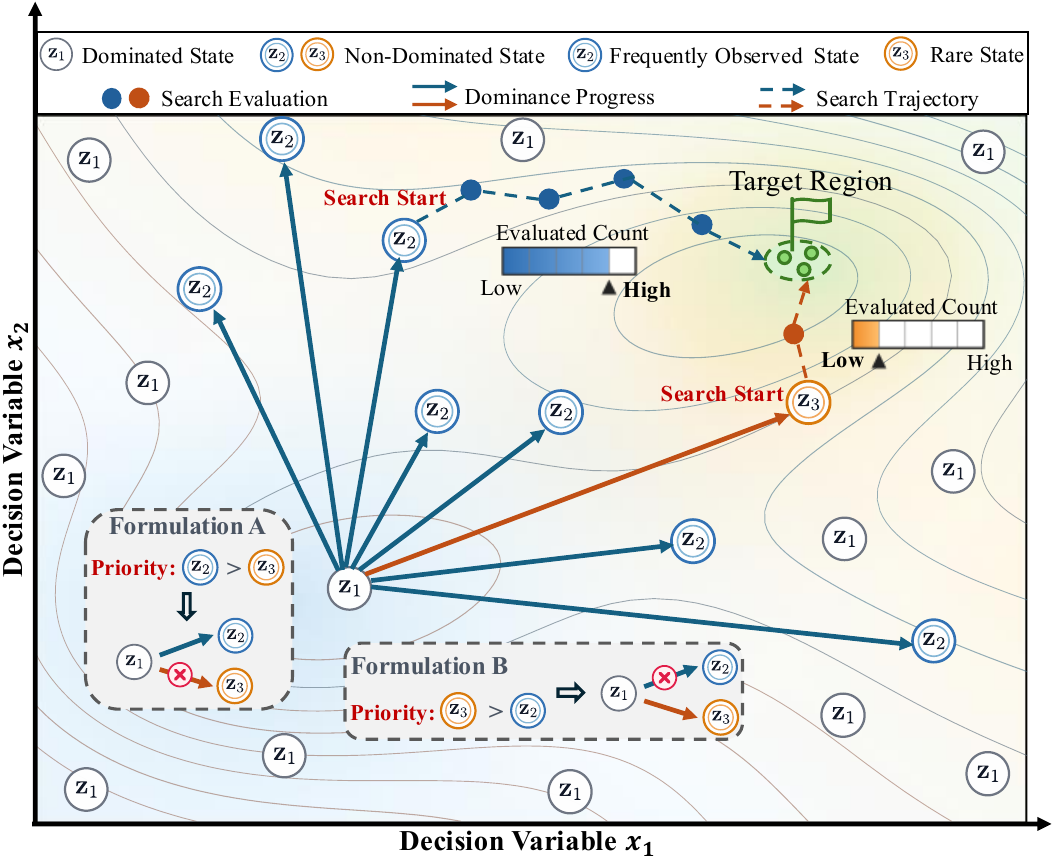}

    \caption{Illustration of search hardness revealed by the initial evaluated data in the decision space.}
    \label{fig2:search path}
\end{figure}

Expensive simulation-driven optimization algorithms usually start from a globally distributed initial set
$\mathcal{D}_{\mathrm{init}}=\{(\mathbf{x}_i^{\mathrm{init}},\mathbf{y}_i^{\mathrm{init}})\}_{i=1}^{N}$, with 
$\mathbf{y}_i^{\mathrm{init}}=\mathcal{S}(\mathbf{x}_i^{\mathrm{init}})$. 
These space-filling evaluations provide sparse observations of the search landscape over the design space, and therefore offer accessible evidence for estimating whether a candidate formulation may induce a lower search cost.

The satisfaction states of the initial samples in $\mathcal{D}_{\mathrm{init}}$ provide partial search trajectories toward $\mathcal{Y}_{\mathcal{R}}^{+}$, as illustrated in Figure~\ref{fig2:search path}.
Specifically, frequently observed states indicate regions that are easily covered by sampling and can be reached by the optimizer without additional guidance, making further prioritization of them less beneficial to the search process.
In contrast, rare states correspond to hard yet promising regions that are difficult for the optimizer to reach naturally, and should be prioritized by the formulation to guide the optimizer into such regions and continue the search.
These observed partial trajectories help estimate which hard trajectories are more likely to reach $\mathcal{Y}_{\mathcal{R}}^{+}$ with fewer evaluations. 
\textbf{This motivates discovering a formulation that prioritizes such hard but promising trajectories, thereby reducing the search cost.}

\subsection{Formulation Search}

\subsubsection{Search Hardness Metric and Anchor Selection}
Based on this finding, we define a search hardness metric for observed satisfaction states and use it to select an anchor state that is hard to reach but promising for further search.
Specifically, given the initial dataset $\mathcal{D}_{\mathrm{init}}$, we use an LLM-based judge to evaluate each response $\mathbf{y}_i$ against the requirement $\mathcal{R}$ and obtain its satisfaction state $\mathbf{z}_i=\mathbf{s}(\mathbf{y}_i)$. 
We denote the observed state set as $\mathcal{Z}_{\mathrm{init}}=\{\mathbf{z}_i\}_{i=1}^{N}$.
The detailed judging prompt is provided in Appendix~\ref{app:state judge prompt template}.

We define the search hardness of each observed state by jointly considering its rarity and progress potential.
For each $\mathbf{z}\in\mathcal{Z}_{\mathrm{init}}$, the search hardness score is defined as
\begin{equation}
h(\mathbf{z})
=
\frac{
\left|
\left\{
i \mid \mathbf{z}_i=\mathbf{z}
\right\}
\right|
}{
\left|
\left\{
i \mid \mathbf{z}\succ \mathbf{z}_i
\right\}
\right|
+\epsilon
},
\end{equation}
where the numerator is the occurrence count of $\mathbf{z}$ in the initial data, with a smaller value indicating higher rarity. 
The denominator measures the progress potential of $\mathbf{z}$ by counting the initial observations whose states are dominated by $\mathbf{z}$. 
Here, $\succ$ follows the Pareto dominance relation~\cite{yuConeConvexityCone1974a} over satisfaction states, where a state dominates another if it satisfies all criteria satisfied by the other state and at least one additional criterion.
The constant $\epsilon$ avoids division by zero.
A lower $h(\mathbf{z})$ therefore indicates a less frequently observed state with stronger dominance support, corresponding to a harder yet more promising state.

Let $\mathcal{Z}_{\mathrm{nd}}$ denote the non-dominated observed state set, defined as
$\mathcal{Z}_{\mathrm{nd}}=
\{\mathbf{z}\in\mathcal{Z}_{\mathrm{init}}
\mid
\nexists\,\mathbf{z}'\in\mathcal{Z}_{\mathrm{init}}:
\mathbf{z}'\succ\mathbf{z}\}$.
The anchor state is then selected from $\mathcal{Z}_{\mathrm{nd}}$ by minimizing $h(\mathbf{z})$:
\begin{equation}
    \mathbf{z}_{\mathrm{anc}}
    =
    \arg\min_{\mathbf{z}\in\mathcal{Z}_{\mathrm{nd}}}
    h(\mathbf{z}).
\end{equation}

\subsubsection{Formulation Search Objective}

For each candidate formulation $\mathcal{F}$, its objectives and constraints induce a feasibility criteria ranking over the initial responses. 
We write $\mathbf{y}_i \triangleright_{\mathcal{F}} \mathbf{y}_j$ if $\mathbf{y}_i$ has higher priority than $\mathbf{y}_j$ under this formulation-induced ranking.
To prioritize the selected anchor state, we construct anchor and other comparison pairs:
\begin{equation}
    \mathcal{P}_{\mathrm{anc}}
    =
    \left\{
    (i,j)
    \mid
    \mathbf{z}_i=\mathbf{z}_{\mathrm{anc}},
    \;
    \mathbf{z}_j\neq\mathbf{z}_{\mathrm{anc}}
    \right\}.
\end{equation}
The formulation search objective is then defined as
\begin{equation}
    \mathcal{F}^{*} 
    =
    \arg\max_{\mathcal{F}\in\widehat{\Phi}_{\mathcal{R}}}
    \frac{1}{|\mathcal{P}_{\mathrm{anc}}|}
    \sum_{(i,j)\in\mathcal{P}_{\mathrm{anc}}}
    \mathbb{I}
    \left[
    \mathbf{y}_i \triangleright_{\mathcal{F}}  \mathbf{y}_j
    \right],
\end{equation}
where $\widehat{\Phi}_{\mathcal{R}}$ denotes the executable and requirement-aligned candidate formulations. 
This objective favors formulations that rank anchor state responses above other responses, thereby guiding the optimizer toward the selected hard but promising search trajectory.

\subsubsection{Formulation Generation and Repair}

In the evolutionary formulation search, each individual is a candidate formulation $\mathcal{F}$ represented as an executable program
$\mathcal{F}(\mathbf{y})=(f_{\mathcal{F}}(\mathbf{y}),\mathbf{g}_{\mathcal{F}}(\mathbf{y}))$,
which maps a simulated response $\mathbf{y}$ to one scalar objective and zero or more constraints.
The objective is minimized, and a response is feasible if
$g_{\mathcal{F},m}(\mathbf{y})\leq 0$ for all $m=1,\ldots,M$.

Candidate formulations are constructed through generation and repair.
A generation LLM $\mathcal{M}_{g}$ first produces a draft formulation
$\tilde{\mathcal{F}}=\mathcal{M}_{g}(\mathcal{R})$.
Since $\tilde{\mathcal{F}}$ may be syntactically executable but semantically misaligned with the design intent, we further employ a repair LLM $\mathcal{M}_{r}$ to check and revise it according to $\mathcal{R}$, yielding $\mathcal{F}=\mathcal{M}_{r}(\tilde{\mathcal{F}},\mathcal{R})$.
This repair step aims to move the draft formulation toward the ideal requirement-aligned set $\Phi_{\mathcal{R}}$ and produces the executable candidate set $\widehat{\Phi}_{\mathcal{R}}$ used in formulation search.
The detailed generation and repair prompts are provided in Appendices~\ref{app:formulation generation} and~\ref{app:formulation repair}, respectively.

\subsubsection{Evolutionary Formulation Refinement}

We search the formulation space by maximizing the formulation search objective defined above.
Starting from a population of repaired formulations, each generation evaluates all candidates on $\mathcal{D}_{\mathrm{init}}$, selects the best-performing formulations as parents, and samples operator prompts to generate offspring through LLM-based variation.

Offspring formulations are generated by applying operator prompts to selected parents.
Rather than relying on a single mutation instruction, we define a set of formulation-space variation operators that modify candidates at different levels of abstraction.
These operators cover component-level recombination, constraint-structure modification, optimization-signal reparameterization, and broader semantic exploration.
The operator set contains four complementary strategies:
\begin{itemize}
    \item \textbf{Component recombination} merges useful objective and constraint components from two best-fitness parents.
    \item \textbf{Constraint restructuring} modifies the number, grouping, or regional organization of constraints.
    \item \textbf{Signal reparameterization} changes how objective values or constraint violations are measured and aggregated.
    \item \textbf{Semantic exploration} introduces larger formulation-level changes to explore alternative formulation families while preserving executability and requirement alignment.
\end{itemize}
The detailed operator prompts are provided in Appendix~\ref{app:operator-prompts}.

Each offspring is repaired by $\mathcal{M}_{r}$ and assigned an objective value using the formulation search objective.
Parents and offspring are then merged and ranked by fitness, and the top candidates are retained as the next population.
The process repeats until the evaluation budget or maximum number of generations is reached, and the formulation with the highest objective value is returned for optimization.

\section{Experiments}

\subsection{Experimental Setup}

\subsubsection{Benchmark Tasks}
We evaluate SHA-PF on a real-world multi-objective calibration problem and five expensive antenna design problems.
We use real-world tasks rather than synthetic benchmark functions because SHA-PF focuses on how different formulations of the same design intent affect the induced optimization process. 
Unlike synthetic benchmarks with fixed objective and constraint definitions, real-world simulation and calibration tasks involve domain-specific performance criteria that can be encoded in multiple valid ways, making formulation-induced search differences more visible and practically relevant.
Detailed descriptions of the HBV and antenna benchmarks are provided in Appendices~\ref{app:hbv benchmark} and~\ref{app:antenna benchmark}, respectively.

\textbf{HBV rainfall-runoff calibration.}
We include the HBV rainfall-runoff model calibration problem as a real-world multi-objective benchmark with a landscape induced by a practical hydrological model~\cite{reedEvolutionaryMultiobjectiveOptimization2013}.
HBV is a conceptual hydrological model with 14 real-valued calibration variables.
The benchmark calibrates the model on the Williams River, West Virginia, using precipitation and streamflow data from the MOPEX dataset.
It involves four hydrological objectives: Nash-Sutcliffe efficiency (NSE) for high-flow accuracy, Box-Cox transformed RMSE (TRMSE) for low-flow accuracy, runoff coefficient error (ROCE) for long-term water balance, and flow-duration-curve slope error (SFDCE) for flow-regime variability.
This task provides a complementary formulation scenario, where multiple conflicting performance criteria may be translated into an optimization problem and different formulations may induce different search behaviors.

\textbf{Antenna design benchmarks.}
We further evaluate SHA-PF on an antenna benchmark suite, which serves as the primary testbed for expensive simulation-driven design.
The suite covers a wide range of antenna structures, from traditional single-layer patches to more intricate multilayer resonator-coupled designs, and includes HAE~\cite{guoDesignWidebandFiltering2024}, CSL~\cite{huDesignCompactSingleLayered2020}, HSL~\cite{liuCompactFilteringPatch2023}, WHG~\cite{yuanWidebandHighGain2022}, and HSE~\cite{liangHighSelectivityHigh2022a}.
These tasks differ in their filtering mechanisms, coupling structures, and radiation characteristics, leading to distinct landscapes and design trade-offs.
In each task, evaluating a candidate design requires electromagnetic simulation, so an inefficient formulation can substantially increase the number of costly simulations needed to reach the target region.
These benchmarks therefore directly test whether SHA-PF can discover formulations that better align the induced search trajectory with the design requirements and guide the optimizer toward requirement-satisfying antenna designs with fewer simulations.

\subsubsection{Baselines}
To the best of our knowledge, no existing method directly addresses search-efficient problem formulation for expensive simulation-driven design. 
We therefore construct baselines representing two common formulation strategies: manual formulation by human experts and direct formulation generation by LLMs.
Specifically, we compare SHA-PF with four formulation baselines, denoted as $\mathcal{F}_{\text{Expert-A}}$, $\mathcal{F}_{\text{Expert-B}}$, $\mathcal{F}_{\text{LLM-A}}$, and $\mathcal{F}_{\text{LLM-B}}$.
Since the same requirement may admit multiple valid formulations, we use two mathematically different formulations for each baseline type to avoid relying on a single baseline formulation.
The expert-based formulations are manually designed and checked by human experts to ensure consistency with the design requirements, while the direct LLM-based formulations are generated using the same requirements and generation prompt as SHA-PF.
The baseline formulation code for the HBV and antenna benchmarks are provided in Appendices~\ref{app:hbv formulations} and~\ref{app:hse formulations}, respectively.
\subsubsection{Implementation Details}

For all benchmark tasks, we use $100$ initial evaluated samples to construct $\mathcal{D}_{\mathrm{init}}$.
To ensure a fair comparison, all LLM-based components use GPT-5-mini, and all formulations are represented as executable Python functions.
We use SCBO~\cite{erikssonScalableGlobalOptimization2019} as the optimization algorithm for all formulations, so that performance differences can be attributed to the formulations rather than the optimizer.
All methods are evaluated under the same budget $B$.
For HBV, the budget is $250$ evaluations.
For each antenna design task, the budget is $300$ HFSS simulations, where HFSS (High Frequency Structure Simulator) is run in ANSYS Electronics Desktop 2018.2.0.
For SHA-PF, each formulation generation strategy produces one candidate formulation.
The population size is set to $P=5$, and the maximum number of evolutionary generations is set to $G_{\max}=25$.
A sensitivity analysis of these parameters is provided in Appendix~\ref{app:parameter-sensitivity}.

\subsection{Main Results}

\begin{figure}[t] 
    \centering
    \includegraphics[width=\linewidth]{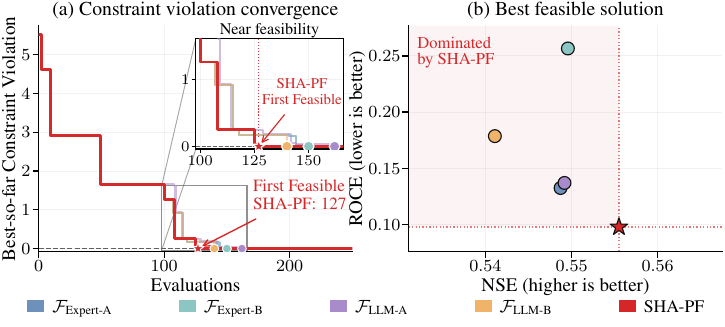}

    \caption{Comparison of feasibility convergence and best feasible solutions on the HBV problem.}
    \label{fig3:hbv_result}
\end{figure}

\subsubsection{HBV Problem}

For the HBV task, the design requirement specifies four threshold criteria:
$\mathrm{NSE}\geq 0.50$, $\mathrm{TRMSE}\leq 1.60$, $\mathrm{ROCE}\leq 0.50$, and $\mathrm{SFDCE}\leq 0.20$.
A solution satisfying all four criteria is treated as feasible.
After feasibility is achieved, higher NSE and lower ROCE are further preferred.

Figure~\ref{fig3:hbv_result} (a) shows the convergence of the best-so-far constraint violation.
SHA-PF reaches feasibility at evaluation $127$, earlier than all compared formulations.
This suggests that its discovered formulation induces a more efficient trajectory toward the feasible region.
The slower reduction of the baselines further shows that valid formulations of the same requirement can still lead to substantially different search efficiency.
Figure~\ref{fig3:hbv_result} (b) compares the best feasible solutions in the NSE-ROCE space.
After feasibility is achieved, higher NSE and lower ROCE are preferred, so the lower-right direction is better.
The solution found by SHA-PF dominates the best solutions obtained by the compared formulations, with higher NSE and lower ROCE.
Thus, SHA-PF not only reaches feasibility earlier but also obtains a better solution under the post-feasibility preference.

Overall, the HBV results show that formulation quality substantially affects search efficiency even with the same optimizer.
This supports the benefit of using hard yet promising states to guide formulation search, which leads to both faster feasibility and better performance.

\subsubsection{Antenna Design Benchmarks}

\begin{figure*}[t] 
    \centering
    \includegraphics[width=\linewidth]{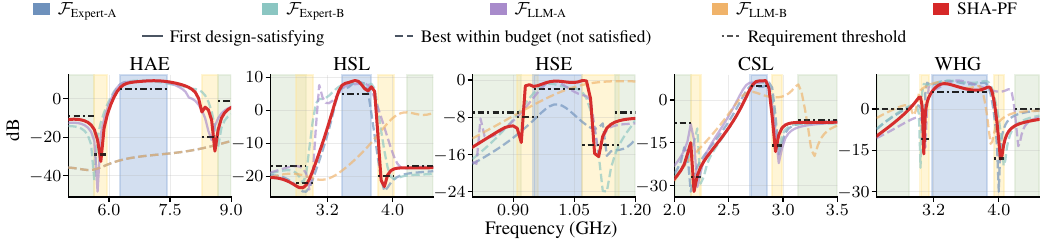}
    \caption{Comparison of formulation methods across five antenna design tasks.}
    \label{fig4:antenna_results}
\end{figure*}

For the antenna design tasks, the design requirements are specified over simulated frequency-response curves.
Each task contains stopband, passband, and null regions.
The stopband regions require the response to be suppressed below given thresholds, the passband regions require the response to remain above given thresholds, and the null regions require radiation nulls.
A design is requirement-satisfying only when all required regions are satisfied.
Detailed task-specific requirements are provided in Appendix~\ref{app:design requirements}.

Table~\ref{tab1:antenna_search_cost} reports the search cost required to find the first requirement-satisfying antenna design, measured by both HFSS simulations and time cost in hours.
SHA-PF succeeds on all five antenna tasks, whereas most expert-based and direct LLM-based formulations fail within the same search budget.
Among the baselines, only $\mathcal{F}_{\text{LLM-A}}$ succeeds on HAE and CSL, requiring $71$ evaluations / $3.1$ hours and $36$ evaluations / $6.0$ hours, respectively.
SHA-PF reaches satisfying designs earlier on both tasks, using $66$ evaluations / $2.7$ hours on HAE and only $10$ evaluations / $2.1$ hours on CSL.
On HSL, HSE, and WHG, all baseline formulations exhaust the evaluation budget without success, whereas SHA-PF finds satisfying designs with \textbf{110--137} fewer simulations and saves \textbf{1.5--11.2} hours.
These results show that SHA-PF provides more reliable and search-efficient formulations across different antenna structures, reducing both HFSS evaluations and search time.

Figure~\ref{fig4:antenna_results} compares the simulated response curves obtained by different formulations.
For successful methods, the figure reports the first requirement-satisfying response; for failed methods, it reports the best response found within the budget.
The baseline formulations often improve parts of the response curve but still violate at least one required region, especially when passband behavior, stopband suppression, and null generation must be satisfied simultaneously.
In contrast, SHA-PF finds response curves that satisfy all required regions across the five tasks.
Overall, the antenna results show that SHA-PF improves the search efficiency of expensive antenna design by finding requirement-satisfying response curves with fewer HFSS simulations and less search time.

\begin{table}[t]
\centering
\footnotesize 
\setlength{\tabcolsep}{2.5pt}
\renewcommand{\arraystretch}{1.2}
\begin{tabular*}{\linewidth}{@{\extracolsep{\fill}}lccccc@{}}
\toprule
\diagbox{Method}{Task} & HAE & HSL & HSE & CSL & WHG \\
\midrule
\multicolumn{6}{@{}c}{\emph{Formulation baselines with SCBO}} \\
\midrule
{$\mathcal{F}_{\text{Expert-A}}$} {\scriptsize (SCBO)}
& --\,/\,13.8 & --\,/\,8.6  & --\,/\,15.1 & --\,/\,31.8 & --\,/\,17.4 \\
{$\mathcal{F}_{\text{Expert-B}}$} {\scriptsize (SCBO)}
& --\,/\,7.8  & --\,/\,14.2 & --\,/\,11.1 & --\,/\,31.2 & --\,/\,19.1 \\
{$\mathcal{F}_{\text{LLM-A}}$} {\scriptsize (SCBO)}
& \underline{71\,/\,3.1} & --\,/\,11.1 & --\,/\,13.5 & 36\,/\,6.0  & --\,/\,16.9 \\
{$\mathcal{F}_{\text{LLM-B}}$} {\scriptsize (SCBO)}
& --\,/\,11.7 & --\,/\,13.6 & --\,/\,12.4 & --\,/\,28.4 & --\,/\,15.9 \\
\midrule
\multicolumn{6}{@{}c}{\emph{SHA-PF with different optimizers}} \\
\midrule
SHA-PF {\scriptsize (SCBO)}
& \textbf{66\,/\,2.7} & \textbf{85\,/\,7.1} & \textbf{63\,/\,3.9} & \underline{10\,/\,2.1} & \textbf{90\,/\,8.6} \\
SHA-PF {\scriptsize (DSI)}
& 85\,/\,7.0 & --\,/\,17.3 & --\,/\,19.2 & \textbf{9\,/\,1.4} & 103\,/\,14.8 \\
SHA-PF {\scriptsize (CEBO)}
& 76\,/\,6.4 & \underline{175\,/\,16.9} & --\,/\,22.6 & 15\,/\,5.2 & \underline{94\,/\,10.6} \\
\bottomrule
\end{tabular*}

\caption{Search cost of different formulation methods to the first requirement-satisfying antenna design, excluding the initial evaluations. Entries are evaluations / hours; best and second-best successful results for each task are shown in bold and underlined, respectively.}
\label{tab1:antenna_search_cost}

\par\smallskip
\begin{minipage}{\linewidth}
\footnotesize
\emph{Note:} ``--'' means that no requirement-satisfying design is found within the search budget. Parentheses indicate the optimizer used for each method. 
\end{minipage}
\end{table}

\subsection{Cross-Optimizer Evaluation}

To examine whether the formulation discovered by SHA-PF is tied to a specific optimizer, we evaluate it with different expensive constrained optimization algorithms across the five antenna design tasks.
Besides SCBO, we include CEBO~\cite{liuConstrainedEvolutionaryBayesian2025} and DSI~\cite{songSurrogateAssistedEvolutionaryFramework2023}.
CEBO is a constrained Bayesian optimization method that models constraints with probabilistic surrogates and incorporates feasibility into the acquisition function.
DSI is a surrogate-assisted evolutionary algorithm for expensive constrained optimization that improves feasibility prediction by sampling near the boundary of the feasible region.
All optimizers use the same SHA-PF formulation and evaluation budget.

Table~\ref{tab1:antenna_search_cost} further reports the cross-optimizer results of SHA-PF.
With the same SHA-PF formulation, SCBO succeeds on all five tasks and achieves the best overall efficiency on most of them.
DSI obtains the lowest cost on CSL but fails on HSL and HSE, while CEBO succeeds on four tasks but fails on HSE and generally requires more evaluations than SCBO.
Nevertheless, SHA-PF with DSI or CEBO still outperforms the formulation baselines in terms of successful tasks, suggesting that optimizer choice is important but not the main reason for the performance gain.
Overall, these results show that SHA-PF is not tied to a specific optimizer.
The discovered formulation provides a more effective search direction across different optimizers, while stronger optimizers such as SCBO can further improve the efficiency of reaching requirement-satisfying antenna designs.

\subsection{Robustness to LLM Backbones}

\begin{figure}[t] 
    \centering
    \includegraphics[width=\linewidth]{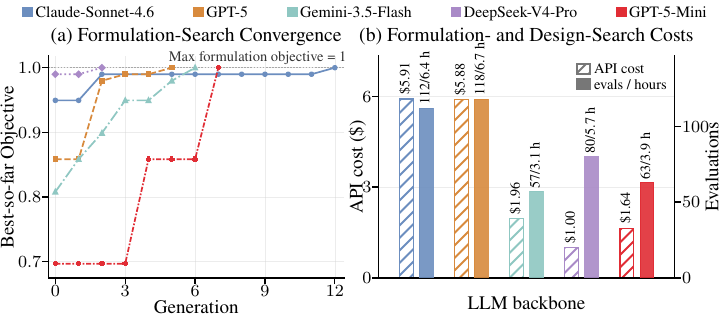}

    \caption{Impact of LLM backbone on formulation search and antenna optimization for the HSE problem.}
    \label{fig6:llm_result}
\end{figure}

We further examine the sensitivity of SHA-PF to the choice of LLM backbone on the HSE task.
Figure~\ref{fig6:llm_result} (a) shows the formulation-search convergence under five different backbones.
Although the initial objective values and convergence speeds vary across models, all backbones eventually reach maximum formulation objective values, suggesting that SHA-PF can refine formulations even when the initial generation quality differs.

Figure~\ref{fig6:llm_result} (b) further compares the cost performance trade-offs of the resulting formulations.
Although all backbones can produce effective formulations after refinement, their design-search costs different.
GPT-5 and Claude-Sonnet-4.6 have higher API costs, but their discovered formulations require more than $100$ evaluations in the antenna design search.
DeepSeek-V4-Pro converges fastest in formulation search and has the lowest API cost, yet its design-search cost remains higher than those of Gemini-3.5-Flash and GPT-5-Mini.
By contrast, Gemini-3.5-Flash and GPT-5-Mini achieve lower design-search costs with relatively low API costs, suggesting that lightweight backbones can provide better overall cost efficiency.

Overall, these results suggest that the LLM backbone affects the convergence speed and cost efficiency of SHA-PF, but it is not the primary factor determining whether SHA-PF can produce an effective formulation.
SHA-PF remains effective across different LLM backbones and does not require the strongest or most expensive model.

\subsection{Ablation Study}

We conduct ablation studies on the antenna design benchmarks to evaluate the contribution of two key components in SHA-PF: anchor selection and formulation repair.
The first variant, w/o Anchor Selection, replaces the proposed anchor selection based on search hardness with random anchor selection.
The second variant, w/o Formulation Repair, removes the repair step after formulation generation.
All variants use the same optimizer, initial data, and evaluation budget as the full SHA-PF.

Table~\ref{tab2:antenna_ablation_results} reports the search cost required to find the first requirement-satisfying antenna design.
Removing anchor selection degrades search efficiency on most tasks.
Although this variant still finds satisfying designs on HAE, HSL, and CSL, it requires more simulations than the full SHA-PF, increasing the evaluation cost from $66$ to $102$, $85$ to $92$, and $10$ to $12$, respectively.
The corresponding time cost increases by about $9.9\%$--$114.8\%$, and the variant fails on HSE and WHG within the budget.
This shows that randomly selected anchors may not correspond to hard yet promising states, and therefore provide weaker guidance for formulation search.

Removing formulation repair causes a more severe degradation.
The w/o Formulation Repair variant fails on all five antenna tasks within the search budget.
This indicates that LLM-generated formulations are often executable but not sufficiently aligned with the design intent.
Without repair, invalid constraint directions, missing requirement regions, or inconsistent objective definitions can remain in the formulation, leading the optimizer toward incorrect search directions.

Overall, the ablation results demonstrate that both components are necessary for SHA-PF.
The anchor selection based on search hardness improves formulation search efficiency, while formulation repair improves the reliability of candidate formulations.
These two components jointly enable SHA-PF to find requirement-satisfying antenna designs using fewer HFSS simulations and less search time.

\begin{table}[t]
\centering
\footnotesize
\setlength{\tabcolsep}{3pt}
\begin{tabular*}{\linewidth}{@{\extracolsep{\fill}}cccc@{}}
\toprule
\raisebox{0.5\normalbaselineskip}{Task} & \shortstack{w/o Anchor\\Selection} & \shortstack{w/o Formulation\\Repair} & \raisebox{0.5\normalbaselineskip}{SHA-PF} \\
\midrule
HAE & 102\,/\,5.8 & --\,/\,7.5 & \textbf{66\,/\,2.7} \\
HSL & 92\,/\,7.8 & --\,/\,13.1 & \textbf{85\,/\,7.1} \\
HSE & --\,/\,11.3 & --\,/\,12.1 & \textbf{63\,/\,3.9} \\
CSL & 12\,/\,3.2 & --\,/\,33.0 & \textbf{10\,/\,2.1} \\
WHG & --\,/\,17.4 & --\,/\,17.7 & \textbf{90\,/\,8.6} \\
\bottomrule
\end{tabular*}
\caption{Search cost of different ablation variants to the first requirement-satisfying antenna design, excluding the initial evaluations. Best results are in bold.}
\label{tab2:antenna_ablation_results}
\par\smallskip
\begin{minipage}{\linewidth}
\footnotesize
\emph{Note:} Entries report evaluations / hours during the search. ``--'' indicates no requirement-satisfying design within the search budget, with hours reported for the full budget.
\end{minipage}
\end{table}

\section{Conclusion}
This work introduces SHA-PF, a search hardness-aware problem formulation framework for expensive simulation-driven design.
Unlike existing LLM-based formulation methods that mainly focus on design-intent alignment, SHA-PF further considers how a formulation affects search efficiency.
It uses initial evaluated data to identify hard yet promising anchor states and searches for executable formulations that prioritize these states through LLM-based generation, repair, and evolutionary refinement.
Experiments on the real-world multi-objective problem and five antenna design benchmarks show that SHA-PF reaches requirement-satisfying designs with fewer evaluations and less search time than expert-based and direct LLM-generated formulations.
Cross-optimizer and ablation studies further show that the discovered formulations remain effective across different optimizers and that both anchor selection and formulation repair contribute to reliable performance.

Despite these results, SHA-PF still depends on the initial evaluated data to estimate search hardness, so anchor selection may be less reliable when these samples are sparse or miss hard yet promising regions.
Future work may explore iterative anchor discovery by using a small number of additional evaluations to update the satisfaction-state structure during formulation search.
The current binary satisfaction state also ignores how close a response is to satisfying each criterion, and incorporating graded satisfaction or violation magnitudes may further improve search hardness estimation.

\bibliography{main.bib}

\FloatBarrier
\clearpage
\setcounter{secnumdepth}{2}
\appendix
\twocolumn[
\begin{center}
{\LARGE\bfseries Appendix}
\end{center}
\vspace{1.5em}
]

\section{Parameter Sensitivity Analysis}
\label{app:parameter-sensitivity}

We investigate the sensitivity of SHA-PF to the population size $P$ and the maximum number of evolutionary generations $G_{\max}$ on the HSE antenna design task.
When varying $P$, we fix $G_{\max}=25$; when varying $G_{\max}$, we fix $P=5$.

\begin{figure}[hbpt]
    \centering
    \includegraphics[width=\linewidth]{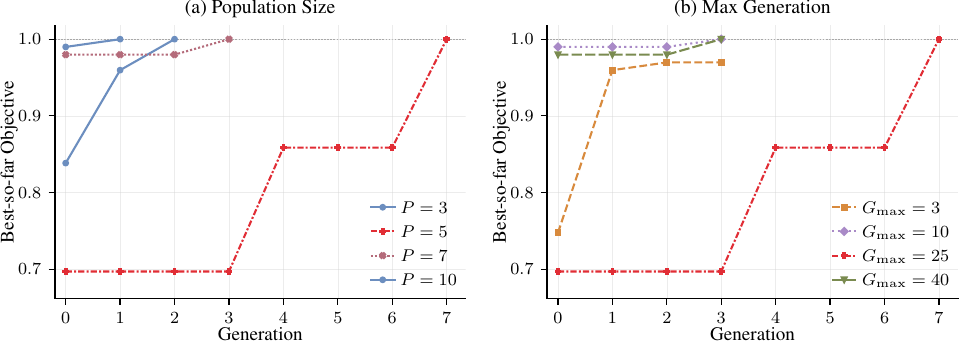}
    \caption{Convergence of the best-so-far formulation search objective under different population sizes and maximum numbers of evolutionary generations on the HSE antenna design task.}
    \label{fig:parameter-sensitivity}
\end{figure}

Figure~\ref{fig:parameter-sensitivity} shows the convergence of the best-so-far objective during formulation search.
As shown in Figure~\ref{fig:parameter-sensitivity}(a), all tested population sizes successfully attain the maximum objective value, while their convergence rates differ because the randomly initialized populations contain candidate formulations of different quality.
Figure~\ref{fig:parameter-sensitivity}(b) further shows that a small $G_{\max}$ can terminate formulation search before the maximum objective value is reached when the initial population is of relatively low quality.
Allowing more generations provides additional opportunities to improve such populations and reach the maximum objective value.

Table~\ref{tab:hse-parameter-sensitivity} evaluates the formulation obtained under each parameter setting in the antenna design search.
The settings $P=5$ and $P=7$ succeed with similar costs, whereas $P=3$ and $P=10$ fail within the simulation budget.
This suggests that a moderate population provides a better balance
between formulation diversity and focused refinement. 
A small population may limit diversity, whereas a large population may make
selection less focused because of the increased number of competing
candidates.
For the generation limit, $G_{\max}=10$ achieves the lowest cost, $G_{\max}=25$ and $40$ also succeed, and $G_{\max}=3$ fails.
Thus, too few generations may terminate refinement prematurely, while additional generations provide no consistent benefit once a strong formulation is found.
We use $P=5$ and $G_{\max}=25$ as the default setting in the main experiments.

\begin{table}[t]
\centering
\footnotesize
\setlength{\tabcolsep}{3pt}
\begin{tabular*}{\linewidth}{@{\extracolsep{\fill}}ccc@{}}
\toprule
Parameter & Value &
\shortstack{Search Cost\\(Evaluations / Hours)} \\
\midrule
\multicolumn{3}{c}{\emph{Population size sensitivity ($G_{\max}=25$)}} \\
\midrule
$P$ & 3  & --\,/\,11.1 \\
$P$ & 5  & 63\,/\,3.9 \\
$P$ & 7  & \textbf{62\,/\,3.8} \\
$P$ & 10 & --\,/\,13.2 \\
\midrule
\multicolumn{3}{c}{\emph{Maximum generation sensitivity ($P=5$)}} \\
\midrule
$G_{\max}$ & 3  & --\,/\,12.4 \\
$G_{\max}$ & 10 & \textbf{46\,/\,3.3} \\
$G_{\max}$ & 25 & 63\,/\,3.9 \\
$G_{\max}$ & 40 & 87\,/\,6.3 \\
\bottomrule
\end{tabular*}
\caption{Parameter sensitivity of SHA-PF on the HSE antenna design task. Best successful result in each group is shown in bold.}
\label{tab:hse-parameter-sensitivity}
\par\smallskip
\begin{minipage}{\linewidth}
\footnotesize
\emph{Note:} $P$ and $G_{\max}$ denote the population size and the maximum number of evolutionary generations, respectively. Entries report evaluations / hours during the search, excluding the initial evaluations. ``--'' indicates that no requirement-satisfying design was found within the budget of 200 evaluations, with hours reported for the full budget.
\end{minipage}
\end{table}

\section{Prompt Template}

This appendix gives the prompt templates used in SHA-PF.
The templates cover satisfaction-state judging, formulation generation, formulation repair, and evolutionary operator prompting. 
Placeholders such as \textit{\{Design\_Requirements\}} and \textit{\{Curve\_Data\}} denote task-specific content supplied at run time.

\subsection{State Judge Prompt Template}
\label{app:state judge prompt template}

The state judge maps a simulated response to a binary satisfaction-state representation.
Given the sampled response curve and the natural-language design requirements, the judge assigns one compliance label to each requirement region. 
These labels form the satisfaction states used for anchor-state selection and formulation evaluation.

\Needspace{0.24\textheight}
\begin{promptcard}[label={prompt:curve-state}]{Response-State Judge Prompt Template}
\begin{promptmetablock}
\begin{promptinput}
Curve Data; Design Requirements; Background Context.
\end{promptinput}

\begin{promptoutput}
JSON Object: Compliance Vector.
\end{promptoutput}
\end{promptmetablock}

\tcblower
\begin{prompttext}

\promptheading{Role}
\begin{promptcontent}
You are an expert evaluator for simulation-derived performance curves.
\end{promptcontent}

\promptheading{Task}
\begin{promptcontent}
\textit{\{Background\_Context\}}. You will be given one performance curve as a list of sampled $(x,y)$ data points in ascending order of $x$. For each numbered region in the design requirements, judge whether the curve meets the stated requirement.
Use binary labels: 1 means that the requirement is met, and 0 means that the requirement is not met.
\end{promptcontent}

\promptheading{Rules}
\begin{promptcontent}
Use only the provided sampled points when making the judgment. Do not interpolate, extrapolate, smooth the curve, or assume unseen values between sampled points. Evaluate each region independently.
For regions that require a local extremum or notch-like behavior, a boundary point may be accepted only if the sampled points support a genuine local feature. If the boundary value is merely part of a monotonic trend continuing into a neighboring region, it should not be treated as a valid local feature.
\end{promptcontent}

\promptheading{Design Requirements}
\begin{promptcontent}
\textit{\{Design\_Requirements\}}
\end{promptcontent}

\promptheading{Curve Data}
\begin{promptcontent}
\textit{\{Curve\_Data\}}
\end{promptcontent}

\promptheading{Output Format}
\begin{promptcontent}
Return a single valid JSON object with exactly the following field:

\noindent\texttt{\{"compliance":[r1,r2,...,rN]\}}

\noindent Each $r_i$ must be either 0 or 1. The array order must match the region numbering. Do not output any text outside the JSON object.
\end{promptcontent}
\end{prompttext}
\end{promptcard}

\subsection{Formulation Generation Prompt Template}\label{app:formulation generation}

The formulation generation prompt produces an executable optimization formulation from the design requirements.
The generated formulation maps a simulated response curve to one minimization objective and zero or more constraints.

\Needspace{0.24\textheight}
\begin{promptcard}[label={prompt:formulation-generation}]{Formulation Generation Prompt Template}
\begin{promptmetablock}
\begin{promptinput}
Background Context; Design Requirements.
\end{promptinput}

\begin{promptoutput}
JSON Object: Objective Function; Constraint Functions.
\end{promptoutput}
\end{promptmetablock}

\tcblower
\begin{prompttext}

\promptheading{Role}
\begin{promptcontent}
You are an expert in simulation-driven engineering optimization and problem formulation.
\end{promptcontent}

\promptheading{Task}
\begin{promptcontent}
Generate one executable optimization formulation for an expensive black-box simulation-driven design problem. Each candidate design is evaluated as a sampled performance curve. The formulation should help an optimizer search efficiently for curves satisfying the design requirements under a limited evaluation budget, rather than merely restating the requirements as direct threshold checks.
\end{promptcontent}

\begin{promptcontent}
For each evaluated curve, the formulation must produce one objective value and zero or more constraint values. Candidate solutions will be ranked by a feasibility-based rule: feasible solutions are preferred over infeasible ones; if both are infeasible, the one with smaller overall constraint violation is preferred; if both are feasible, the one with the better objective value is preferred.
\end{promptcontent}

\promptheading{Guidance}
\begin{promptcontent}
The design requirements define final success, but the formulation should provide smoother and more informative search guidance. Prefer graded quality or violation signals that distinguish partially improved curves. Avoid relying only on hard binary checks or trivial one-to-one restatements of the acceptance rules when continuous signals are possible.
\end{promptcontent}

\promptheading{Formulation Rules}
\begin{promptcontent}
Use a unified minimization convention for objectives. Express every constraint in the standard form $g(\cdot)\le 0$, where a constraint is satisfied when its returned value is less than or equal to zero. Use exactly one objective function named \texttt{obj1}. The number of constraint functions may be zero, one, or more. A formulation may use composite constraints that jointly represent multiple design intentions, as long as it remains consistent with the requirements and provides useful optimization guidance.
\end{promptcontent}

\promptheading{Design Context}
\begin{promptcontent}
\textit{\{Background\_Context\}}
\end{promptcontent}

\promptheading{Design Requirements}
\begin{promptcontent}
\textit{\{Design\_Requirements\}}
\end{promptcontent}

\promptheading{Output Format}
\begin{promptcontent}
Return exactly one valid JSON object with keys \texttt{"objectives"} and \texttt{"constraints"}. The \texttt{"objectives"} field must be an array containing one executable Python function string. The objective function must use the signature \texttt{def obj1(curve): ...}, take \texttt{curve} as a 2D NumPy array whose first column is the independent variable and second column is the performance value, and return a scalar float to be minimized.
The \texttt{"constraints"} field must be an array of executable Python function strings. Each constraint function must be named \texttt{cK}, use the signature \texttt{def cK(curve): ...}, take the same \texttt{curve} input, and return a scalar float such that the constraint is satisfied when the value is less than or equal to zero. Use NumPy only, and do not output any text outside the JSON object.
\end{promptcontent}
\end{prompttext}
\end{promptcard}

\subsection{Formulation Repair Prompt Template}\label{app:formulation repair}

The repair prompt checks a generated formulation before evaluation.
It targets clear logical errors and mismatches with the design requirements. When such a problem is found, the LLM makes a minimal local edit while preserving the original formulation structure whenever possible.

\Needspace{0.24\textheight}
\begin{promptcard}[label={prompt:formulation-repair}]{Formulation Repair Prompt Template}
\begin{promptmetablock}
\begin{promptinput}
Design Requirements; Candidate Formulation.
\end{promptinput}

\begin{promptoutput}
JSON Object: Repair Summary; Repaired Formulation.
\end{promptoutput}
\end{promptmetablock}

\tcblower
\begin{prompttext}

\promptheading{Role}
\begin{promptcontent}
You are a checker and repairer for optimization formulations used in expensive black-box simulation optimization.
\end{promptcontent}

\promptheading{Task}
\begin{promptcontent}
Inspect the given formulation and repair it only when there is a clear logical error or a clear mismatch with the design intent. This is a repair task, not a reformulation or performance-improvement task. Do not rewrite, simplify, or redesign the formulation merely because another version may look cleaner, simpler, or potentially better.
\end{promptcontent}

\begin{promptcontent}
The formulation contains exactly one objective function and zero or more constraint functions. Its purpose is to provide meaningful search guidance during optimization while remaining aligned with the design requirements. The repaired formulation should remain an optimization-guiding formulation rather than a direct restatement of the acceptance rules.
\end{promptcontent}

\promptheading{Checks}
\begin{promptcontent}
Only check for clear sign-direction errors, clear mismatch between a term and the stated design intent, clear logical conflict between the objective and constraints, duplicated encouragement or duplicated penalty that creates logical imbalance, missing handling of important empty-mask, boundary, or abnormal cases, or formulation logic that would clearly mislead the optimizer away from the intended behavior.
\end{promptcontent}

\promptheading{Repair Rules}
\begin{promptcontent}
When repair is necessary, preserve the original structure as much as possible. Keep exactly one objective function, and keep the same number of constraints unless one is clearly invalid and cannot be minimally repaired. Prefer minimal local edits, such as adjusting signs, thresholds, local terms, local aggregations, or boundary handling, rather than rewriting full functions. Do not modify unaffected parts. If no clear logical problem is found, return the original formulation unchanged. When in doubt, keep the original formulation unchanged.
\end{promptcontent}

\promptheading{Formulation Conventions}
\begin{promptcontent}
The objective function is defined for minimization, so lower objective values must indicate better solutions. Each constraint must follow the convention $g(\texttt{curve})\le 0$ means satisfied and $g(\texttt{curve})>0$ means violated. The repaired formulation must preserve these conventions and must not reverse the optimization direction.
\end{promptcontent}

\promptheading{Design Requirements}
\begin{promptcontent}
\textit{\{Design\_Requirements\}}
\end{promptcontent}

\promptheading{Formulation to Inspect}
\begin{promptcontent}
\textit{\{Candidate\_Formulation\}}
\end{promptcontent}

\promptheading{Output Format}
\begin{promptcontent}
Return exactly one valid JSON object with keys \textbf{repair summary} and \textbf{repaired formulation}. The repair summary must briefly state whether any clear logical problem was found and, if so, what was minimally repaired. If no clear problem was found, explicitly state that the original formulation was kept unchanged.
\end{promptcontent}

\begin{promptcontent}
The repaired formulation must be a JSON object with keys \textbf{objectives} and \textbf{constraints}. The objectives field must contain exactly one executable Python function string, and the constraints field must contain an array of executable Python function strings. Do not output any text outside the JSON object.
\end{promptcontent}
\end{prompttext}
\end{promptcard}

\subsection{Operator Prompts}
\label{app:operator-prompts}

The operator prompts below guide evolutionary formulation refinement. 
Each operator is inserted into the offspring-generation prompt together with selected parent formulations. 
We list the operators separately because they only specify how a new candidate should differ from its parent formulation(s); the shared context and output format follow the formulation generation template.

\Needspace{0.22\textheight}
\begin{operatorpromptbox}
\operatorprompt{Component Recombination}
{Create one new formulation by recombining the most useful parts of the two parent formulations. Keep the components that already provide effective optimization guidance, and combine them into a stronger formulation rather than making only superficial changes.}

\operatorprompt{Constraint Restructuring}
{Create one structurally different new formulation based on the parent formulation. Actively explore a different constraint organization, such as using fewer constraints, more constraints, or composite constraints across multiple regions. Do not limit yourself to the parent's current constraint structure.}

\operatorprompt{Signal Reparameterization}
{Create one new formulation by exploring a different way to define the objective signal and/or constraint-violation signal from the parent formulation. You may change how objective values or constraint violations are measured, aggregated, or combined, as long as the formulation remains aligned with the design intent and provides useful optimization guidance.}

\operatorprompt{Semantic Exploration}
{Create one new formulation that is intentionally more different from the parent formulation. You may substantially change the objective definition, the number of constraints, or the way regional design intentions are organized, as long as the resulting formulation remains executable and aligned with the design requirements. The goal is to explore a new formulation family rather than making only small local edits.}
\end{operatorpromptbox}

\section{Benchmark Problems}

We evaluate SHA-PF on the real-world multi-objective benchmark and five antenna design tasks.

\subsection{HBV Rainfall--Runoff Benchmark}\label{app:hbv benchmark}

HBV is a lumped conceptual rainfall--runoff model used here as a real-world multi-objective calibration benchmark~\cite{reedEvolutionaryMultiobjectiveOptimization2013}. 
The benchmark calibrates the model for the Williams River, West Virginia, United States (USGS Gage 03186500), using precipitation and streamflow records from the MOPEX dataset.
The native benchmark simulates an 11-year period from October 1, 1961 to September 30, 1972, where the first year is used as warmup and the remaining 10 years are used for objective evaluation.
Each candidate solution is a 14-dimensional real-valued parameter vector, and evaluating it runs the HBV simulator to produce modeled daily streamflow.

\begin{table}[t]
\centering
\footnotesize
\setlength{\tabcolsep}{4pt}
\begin{tabular}{M{0.32\columnwidth}M{0.58\columnwidth}}
\toprule
Item & Description \\
\midrule
Task & HBV rainfall--runoff model calibration \\
Site and data & Williams River, West Virginia; MOPEX precipitation and streamflow records \\
Calibration window & 10 post-warmup years within the 1961--1972 simulation period \\
Problem size & 14 continuous variables and 4 objectives \\
Evaluation budget & 250 simulator evaluations \\
\bottomrule
\end{tabular}
\caption{Summary of the HBV benchmark setting.}
\label{tab:hbv-summary}
\end{table}

Let $\mathbf{x}\in\mathcal{X}\subset\mathbb{R}^{14}$ denote an HBV calibration vector.
The feasible parameter domain is $\mathcal{X}=[\ell_1,u_1]\times\cdots\times[\ell_{14},u_{14}]$ shown in Table~\ref{tab:hbv-variables}. 
For each $\mathbf{x}$, the simulator returns a modeled daily streamflow sequence $\hat{q}_{1:N}(\mathbf{x})$, which is compared with the observed streamflow sequence $q_{1:N}$ and precipitation sequence $p_{1:N}$ after the warmup period.

\begin{table}[t]
\centering
\footnotesize
\setlength{\tabcolsep}{3pt}
\begin{tabular}{M{0.22\columnwidth}M{0.19\columnwidth}M{0.47\columnwidth}}
\toprule
Variable & Range & Role \\
\midrule
$L$ (mm) & $[0,100]$ & Response-routine storage threshold \\
$K_0$ (d) & $[0.5,20]$ & Fast reservoir recession time scale \\
$K_1$ (d) & $[1,100]$ & Intermediate reservoir recession time scale \\
$K_2$ (d) & $[10,20000]$ & Slow reservoir recession time scale \\
Perc (mm/d) & $[0,100]$ & Percolation rate \\
LP & $[0.3,1]$ & Evapotranspiration reduction limit \\
Fcap (mm) & $[0,2000]$ & Soil moisture storage capacity \\
$B$ & $[0,7]$ & Soil moisture shape parameter \\
MaxBas (d) & $[24,120]$ & Routing/base length parameter \\
TT (${}^\circ$C) & $[-3,3]$ & Temperature threshold for snow/rain partitioning \\
DDF & $[0,20]$ & Degree-day snowmelt factor \\
CFR & $[0,1]$ & Refreezing coefficient \\
CWH & $[0,0.8]$ & Liquid-water holding capacity in snow \\
TTI (${}^\circ$C) & $[0,7]$ & Temperature transition interval \\
\bottomrule
\end{tabular}
\caption{Decision variables and bounds in the HBV benchmark.}
\label{tab:hbv-variables}
\end{table}

Table~\ref{tab:hbv-objectives} summarizes the four criteria used in the original benchmark.
\begin{table}[t]
\centering
\footnotesize
\setlength{\tabcolsep}{3pt}
\begin{tabular}{M{0.16\columnwidth}M{0.18\columnwidth}M{0.56\columnwidth}}
\toprule
Objective & Direction & Hydrological meaning \\
\midrule
NSE & Maximize & Nash--Sutcliffe efficiency; emphasizes high-flow accuracy \\
TRMSE & Minimize & RMSE after a Box--Cox transformation with exponent $0.3$; emphasizes low-flow accuracy \\
ROCE & Minimize & Absolute runoff coefficient error averaged over calibration years; measures long-term water balance \\
SFDCE & Minimize & Absolute percent error in the middle-slope segment of the flow-duration curve; measures flow variability \\
\bottomrule
\end{tabular}
\caption{Objectives in the HBV calibration benchmark.}
\label{tab:hbv-objectives}
\end{table}
More explicitly, using $\hat{q}_t$ for simulated flow, $q_t$ for observed flow, and $Y$ for the number of post-warmup calibration years, the implementation computes TRMSE by applying the Box--Cox map $b(z)=(z^{0.3}-1)/0.3$ to daily flows:
\begin{equation}
\mathrm{TRMSE}(\mathbf{x})=
\left(\frac{1}{N}\sum_{t=1}^{N}
\left[b(\hat{q}_t(\mathbf{x}))-b(q_t)\right]^2\right)^{1/2}.
\end{equation}
The runoff coefficient error is computed from annual mean simulated flow, observed flow, and precipitation:
\begin{equation}
\mathrm{ROCE}(\mathbf{x})=
\left|
\frac{1}{Y}\sum_{y=1}^{Y}
\frac{\bar{\hat{q}}_y(\mathbf{x})-\bar{q}_y}{\bar{p}_y}
\right|.
\end{equation}
The flow-duration-curve slope error compares the middle-slope segment of the simulated and observed sorted flow curves:
\begin{equation}
\mathrm{SFDCE}(\mathbf{x})=
\left|
\frac{
(\hat{q}_{0.67}-\hat{q}_{0.33})-(q_{0.67}-q_{0.33})
}{
q_{0.67}-q_{0.33}
}
\right|,
\end{equation}
where $q_{\alpha}$ and $\hat{q}_{\alpha}$ denote the corresponding empirical flow-duration quantiles.
The NSE objective follows the standard efficiency form:
\begin{equation}
\mathrm{NSE}(\mathbf{x})=
1-\frac{\sum_{t=1}^{N}(\hat{q}_t(\mathbf{x})-q_t)^2}
{\sum_{t=1}^{N}(q_t-\bar{q})^2}.
\end{equation}

The design requirement for this benchmark are given in Section~\ref{app:design requirements}.

\subsection{Antenna Design Benchmarks}
\label{app:antenna benchmark}

The antenna suite contains five filtering patch antenna design tasks: HAE~\cite{guoDesignWidebandFiltering2024}, CSL~\cite{huDesignCompactSingleLayered2020}, HSL~\cite{liuCompactFilteringPatch2023}, WHG~\cite{yuanWidebandHighGain2022}, and HSE~\cite{liangHighSelectivityHigh2022a}.
For each task, a candidate solution is a vector of continuous geometric parameters in millimeters.
Instantiating this vector in the antenna model defines one HFSS simulation, and the simulator returns a sampled frequency-domain response over the task-specific sweep range.

All antenna evaluations are performed in ANSYS Electronics Desktop HFSS 2018.2 using a common adaptive solution configuration.
As shown in Table~\ref{tab:hfss-settings}, the adaptive solver is limited to 15 passes with a maximum $\Delta S$ tolerance of 0.02.
The frequency-sweep sample count specifies the number of response points returned by the simulator for each antenna; these sampled responses are the curves used by the requirement state stage.

\begin{table}[H]
\centering
\footnotesize
\setlength{\tabcolsep}{2pt}
\begin{tabular}{C{0.22\columnwidth}C{0.16\columnwidth}C{0.10\columnwidth}C{0.10\columnwidth}C{0.10\columnwidth}C{0.10\columnwidth}C{0.10\columnwidth}}
\toprule
\multicolumn{2}{c}{Adaptive solution} & \multicolumn{5}{c}{Frequency sweep samples} \\
\cmidrule(lr){1-2}\cmidrule(lr){3-7}
Max passes & Max $\Delta S$ & HAE & CSL & HSL & WHG & HSE \\
\midrule
15 & 0.02 & 51 & 51 & 51 & 101 & 41 \\
\bottomrule
\end{tabular}
\caption{HFSS solution settings for the antenna simulations.}
\label{tab:hfss-settings}
\end{table}

\begin{table*}[t]
\centering
\footnotesize
\setlength{\tabcolsep}{4pt}
\begin{tabular}{C{0.08\textwidth}C{0.20\textwidth}C{0.13\textwidth}C{0.10\textwidth}C{0.13\textwidth}C{0.13\textwidth}C{0.12\textwidth}}
\toprule
Antenna & Structure & Full frequency range (GHz) & Dim. & Center freq. (GHz) & Bandwidth (\%) & Avg. eval. time (s) \\
\midrule
HAE & 1 layer; $0.68\times1.14\,\lambda_0^2$ & 5.00--9.00 & 14 & 6.85 & 17.0 & 532 \\
CSL & 1 layer; $0.74\times0.45\,\lambda_0^2$ & 2.00--3.50 & 13 & 2.77 & 5.1 & 963 \\
HSL & 2 layers; $0.38\times0.38\,\lambda_0^2$ & 2.50--4.50 & 19 & 3.50 & 9.7 & 713 \\
WHG & 3 layers; $0.86\times0.86\,\lambda_0^2$ & 2.50--4.50 & 15 & 3.48 & 18.8 & 853 \\
HSE & 3 layers; $1.00\times1.00\,\lambda_0^2$ & 0.80--1.20 & 23 & 1.00 & 16.0 & 511 \\
\bottomrule
\end{tabular}
\caption{Summary of the five antenna design benchmarks.}
\label{tab:antenna-summary}
\end{table*}

Table~\ref{tab:antenna-summary} lists the structure, dimensionality, sweep range, and average HFSS evaluation time for each antenna.
The tasks contain 13--23 design variables, with average simulation times ranging from 511 to 963 seconds.

Each run uses a budget of 300 HFSS evaluations.
The feasible design domain is the Cartesian product of the per-parameter intervals in Table~\ref{tab:antenna-search-space}.
All candidate geometries submitted to HFSS must lie within these bounds.

\begin{table*}[t]
\centering
\tiny
\setlength{\tabcolsep}{2.8pt}
\renewcommand{\arraystretch}{1.08}
\begin{tabular*}{\textwidth}{@{\extracolsep{\fill}}l*{14}{c}@{}}
\toprule
\textbf{HAE} & $l_g$ & $w_g$ & $w_s$ & $l_s$ & $d_1$ & $d$ & $d_2$ & $d_3$ & $w_{s1}$ & $l_{s1}$ & $w_t$ & $l_t$ & $l_{t1}$ & $h$ \\
\midrule
\textbf{Min} & 49 & 20 & 0.2 & 4 & 2.0 & 2 & 4.0 & 3.00 & 0.2 & 2 & 0.3 & 0.8 & 4.5 & 1.5 \\
\textbf{Max} & 60 & 40 & 1.0 & 8 & 5.0 & 5 & 11.0 & 10.00 & 1.0 & 6 & 0.7 & 1.2 & 6.5 & 2.5 \\
\bottomrule
\end{tabular*}

\vspace{2pt}
\begin{tabular*}{\textwidth}{@{\extracolsep{\fill}}l*{13}{c}@{}}
\toprule
\textbf{CSL} & $g$ & $a$ & $f_a$ & $f$ & $m_w$ & $m_l$ & $d$ & $d_1$ & $W_p$ & $L_p$ & $w_w$ & $l_l$ & $L_s$ \\
\midrule
\textbf{Min} & 0.6 & 0.5 & 3 & 8 & 1 & 5.0 & 4 & 13 & 30 & 30 & 6 & 1.5 & 1.8 \\
\textbf{Max} & 2.0 & 1.5 & 15 & 15 & 4 & 12.0 & 12 & 16 & 40 & 36 & 10 & 2.5 & 2.2 \\
\bottomrule
\end{tabular*}

\vspace{2pt}
\begin{tabular*}{\textwidth}{@{\extracolsep{\fill}}l*{19}{c}@{}}
\toprule
\textbf{HSL} & $w_g$ & $l_g$ & $w$ & $l$ & $pw$ & $pl$ & $w_s$ & $l_{s1}$ & $w_{s2}$ & $l_{s2}$ & $w_1$ & $l_1$ & $w_2$ & $l_2$ & $l_{i1}$ & $l_{i2}$ & $d$ & $h$ & $hh$ \\
\midrule
\textbf{Min} & 60 & 60 & 40 & 40 & 25.0 & 25.0 & 2.2 & 20.0 & 2.2 & 20.0 & 1.0 & 5.0 & 0.3 & 5.0 & 2.0 & 2.0 & 8 & 2.5 & 7.0 \\
\textbf{Max} & 70 & 70 & 50 & 50 & 35.0 & 35.0 & 4.0 & 40.0 & 4.0 & 40.0 & 2.0 & 10.0 & 1.0 & 10.0 & 4.0 & 4.0 & 15 & 3.5 & 8.0 \\
\bottomrule
\end{tabular*}

\vspace{2pt}
\begin{tabular*}{\textwidth}{@{\extracolsep{\fill}}l*{15}{c}@{}}
\toprule
\textbf{WHG} & $h$ & $h_a$ & $h_b$ & $l_g$ & $w_p$ & $l_{p1}$ & $l_{p2}$ & $l_{p3}$ & $l_f$ & $w_f$ & $d_f$ & $l_s$ & $w_s$ & $d$ & $g$ \\
\midrule
\textbf{Min} & 0.6 & 2 & 2.0 & 90 & 7.5 & 26 & 44 & 20 & 10.0 & 1.0 & 5 & 12 & 1.5 & 1.8 & 0.9 \\
\textbf{Max} & 1.4 & 4 & 3.5 & 110 & 8.5 & 30 & 46 & 32 & 30.0 & 3.0 & 15 & 20 & 2.5 & 2.2 & 1.1 \\
\bottomrule
\end{tabular*}

\vspace{2pt}
\begin{tabular*}{\textwidth}{@{\extracolsep{\fill}}l*{23}{c}@{}}
\toprule
\textbf{HSE} & $l_g$ & $d$ & $d_p$ & $h_f$ & $w_{fp}$ & $l_{fp}$ & $d_{po}$ & $t_{fp}$ & $l_{p1}$ & $g_f$ & $t$ & $l_{p2}$ & $t_m$ & $g_{p1}$ & $g_{p2}$ & $l_{p3}$ & $l_{s1}$ & $w_{s1}$ & $l_{s2}$ & $w_{s2}$ & $l_{s3}$ & $w_{s3}$ & $k$ \\
\midrule
\textbf{Min} & 280 & 80 & 0.5 & 4.0 & 4 & 20 & 1.5 & 0.2 & 80 & 1.0 & 0.3 & 80 & 0.05 & 4 & 4.0 & 80 & 10 & 1 & 30 & 1 & 30 & 1.0 & 0.10 \\
\textbf{Max} & 320 & 120 & 1.5 & 8.0 & 12 & 32 & 3.0 & 0.8 & 120 & 2.0 & 0.8 & 120 & 1.00 & 12 & 12.0 & 120 & 50 & 5 & 80 & 5 & 80 & 5.0 & 0.30 \\
\bottomrule
\end{tabular*}
\caption{Search-space settings for the antenna design variables. All values are in millimeters.}
\label{tab:antenna-search-space}
\end{table*}

\subsection{Design Requirements}
\label{app:design requirements}

We use the following design requirements to judge whether a simulated response satisfies the target behavior.
The HBV requirements specify hydrological thresholds and post-feasibility preferences. 
The antenna requirements specify the target filtering response through frequency regions and threshold values.

\begin{requirementcard}[label={req:hbv-design}]{HBV Design Requirement Specification}
\reqheading{Design Intent}
The calibrated HBV parameter setting must satisfy all four hydrological thresholds.
NSE is a maximization metric and should be at or above its threshold, whereas TRMSE, ROCE, and SFDCE are minimization metrics and should each be at or below their thresholds.
After all thresholds are satisfied, the formulation should prefer samples with larger NSE and smaller ROCE.
TRMSE and SFDCE remain secondary tie-breaking signals once they are within their acceptable ranges.

\reqheading{Final Acceptance Criteria}
\reqitem{Criterion 1}{NSE}{$m_0$ must satisfy $m_0\geq0.50$.}
\reqitem{Criterion 2}{TRMSE}{$m_1$ must satisfy $m_1\leq1.60$.}
\reqitem{Criterion 3}{ROCE}{$m_2$ must satisfy $m_2\leq0.50$.}
\reqitem{Criterion 4}{SFDCE}{$m_3$ must satisfy $m_3\leq0.20$.}

\end{requirementcard}

The five antenna tasks use the same requirement template, with task-specific frequency intervals and dB thresholds.
Requirement~\ref{req:antenna-design} gives the common template, and Table~\ref{tab:antenna-requirements} reports the task-specific thresholds.

\begin{requirementcard}[label={req:antenna-design}]{Antenna Design Requirement Template}
\reqheading{Design Intent}
The engineer expects the final optimized curve to exhibit a band-pass-like response with two notch regions and suppressed responses on both sides.
Specifically, the response should remain low in the low-frequency stopband, form a clear notch in the low-frequency null region, stay high across the central passband, form a deeper notch in the high-frequency null region, and remain suppressed again in the high-frequency stopband.
Overall, the desired curve should progressively approach a stopband-null-passband-null-stopband shape.

\reqheading{Final Acceptance Criteria}
\reqitem{Region 1}{Low-frequency stopband}{All sampled values in $[\ell_{\mathrm{LS}},u_{\mathrm{LS}}]$ must be no higher than $\tau_{\mathrm{LS}}$.}
\reqitem{Region 2}{Low-frequency null region}{The sampled curve in $[\ell_{\mathrm{LN}},u_{\mathrm{LN}}]$ must contain a distinct local minimum whose value is no higher than $\tau_{\mathrm{LN}}$.}
\reqitem{Region 3}{Passband}{All sampled values in $[\ell_{\mathrm{PB}},u_{\mathrm{PB}}]$ must be at least $\tau_{\mathrm{PB}}$.}
\reqitem{Region 4}{High-frequency null region}{The sampled curve in $[\ell_{\mathrm{HN}},u_{\mathrm{HN}}]$ must contain a distinct local minimum whose value is no higher than $\tau_{\mathrm{HN}}$.}
\reqitem{Region 5}{High-frequency stopband}{All sampled values in $[\ell_{\mathrm{HS}},u_{\mathrm{HS}}]$ must be no higher than $\tau_{\mathrm{HS}}$.}

\end{requirementcard}

\begin{table*}[t]
\centering
\scriptsize
\setlength{\tabcolsep}{2pt}
\renewcommand{\arraystretch}{1.08}
\begin{tabular}{C{0.07\textwidth}C{0.17\textwidth}C{0.17\textwidth}C{0.16\textwidth}C{0.17\textwidth}C{0.17\textwidth}}
\toprule
Antenna & LS: $[\ell_{\mathrm{LS}},u_{\mathrm{LS}}]$; $\tau_{\mathrm{LS}}$ & LN: $[\ell_{\mathrm{LN}},u_{\mathrm{LN}}]$; $\tau_{\mathrm{LN}}$ & PB: $[\ell_{\mathrm{PB}},u_{\mathrm{PB}}]$; $\tau_{\mathrm{PB}}$ & HN: $[\ell_{\mathrm{HN}},u_{\mathrm{HN}}]$; $\tau_{\mathrm{HN}}$ & HS: $[\ell_{\mathrm{HS}},u_{\mathrm{HS}}]$; $\tau_{\mathrm{HS}}$ \\
\midrule
HAE & $[5.00,5.64]$; $-9$ & $[5.64,5.96]$; $-29$ & $[6.27,7.43]$; $5$ & $[8.28,8.68]$; $-20$ & $[8.68,9.00]$; $-1$ \\
CSL & $[2.00,2.15]$; $-8$ & $[2.15,2.24]$; $-27$ & $[2.70,2.85]$; $5$ & $[2.90,2.99]$; $-16$ & $[3.14,3.50]$; $-7$ \\
HSL & $[2.50,2.94]$; $-17$ & $[2.82,3.02]$; $-22$ & $[3.38,3.74]$; $5$ & $[3.82,4.02]$; $-20$ & $[4.18,4.50]$; $-17$ \\
WHG & $[2.50,2.90]$; $0$ & $[3.04,3.14]$; $-11$ & $[3.18,3.86]$; $6$ & $[3.94,4.06]$; $-18$ & $[4.20,4.50]$; $0$ \\
HSE & $[0.80,0.92]$; $-7$ & $[0.91,0.96]$; $-8$ & $[0.95,1.07]$; $-2$ & $[1.07,1.16]$; $-14$ & $[1.15,1.20]$; $-7$ \\
\bottomrule
\end{tabular}
\caption{Task-specific antenna requirement intervals and thresholds. Intervals are in GHz and thresholds are in dB.}
\label{tab:antenna-requirements}
\end{table*}

\section{Formulations}

We provide the executable formulation functions used by SHA-PF and all formulation baselines.
For readability, comments are omitted; each block contains only the code supplied to the optimizer.
All formulations follow the same interface and optimization convention: each function maps the simulator response to one scalar objective and zero or more scalar constraints, objectives are minimized, and a constraint is satisfied when its returned value is no greater than zero.

We construct four formulation baselines. Expert-A and Expert-B are independently developed by two domain experts from the same design requirements. For the LLM baselines, GPT-5-mini is invoked 20 times using the same task information and prompt provided in Section~\ref{app:formulation generation}. All generated formulations are manually inspected to exclude invalid or requirement-inconsistent formulations, after which the two experts independently select the formulation they consider most promising, yielding LLM-A and LLM-B. For all methods, the simulator, initial evaluations, SCBO implementation, evaluation budget, and success test are fixed, such that only the formulation supplied to the optimizer differs.

\subsection{HBV Formulations}
\label{app:hbv formulations}

The HBV formulations operate on the four metrics ordered as NSE, TRMSE, ROCE, and SFDCE.

\formulationsubheading{SHA-PF}
\formulationtitle[form:hbv-shapf]{HBV SHA-PF}
\begin{formulationcode}
def obj1(metrics):
    NSE = float(metrics[0])
    TRMSE = float(metrics[1])
    ROCE = float(metrics[2])
    SFDCE = float(metrics[3])
    t_NSE = 0.5
    t_TRMSE = 1.6
    t_ROCE = 0.5
    t_SFDCE = 0.2
    short_NSE = max(0.0, t_NSE - NSE)
    excess_TRMSE = max(0.0, TRMSE - t_TRMSE)
    excess_ROCE = max(0.0, ROCE - t_ROCE)
    excess_SFDCE = max(0.0, SFDCE - t_SFDCE)
    scale_NSE = 1.0
    scale_TRMSE = 2.0
    scale_ROCE = 1.0
    scale_SFDCE = 1.0
    pen_NSE = (short_NSE / scale_NSE) ** 2
    pen_TRMSE = (excess_TRMSE / scale_TRMSE) ** 2
    pen_ROCE = (excess_ROCE / scale_ROCE) ** 2
    pen_SFDCE = (excess_SFDCE / scale_SFDCE) ** 2
    infeasibility_penalty = 100.0 * (pen_NSE + pen_TRMSE + pen_ROCE + pen_SFDCE)
    feasible = short_NSE <= 0.0 and excess_TRMSE <= 0.0 and (excess_ROCE <= 0.0) and (excess_SFDCE <= 0.0)
    if not feasible:
        tie_break = -5.0 * max(0.0, min(1.0, NSE)) + 1.0 * TRMSE + 1.0 * ROCE + 0.5 * SFDCE
        return infeasibility_penalty + tie_break
    else:
        effective_NSE = min(NSE, 0.7)
        if NSE < 0.7:
            w1 = 20.0
            w2 = 10.0
        else:
            w1 = 10.0
            w2 = 20.0
        w3 = 2.0
        w4 = 2.0
        obj_value = -w1 * effective_NSE + w2 * ROCE + w3 * TRMSE + w4 * SFDCE
        obj_value += 1e-06 * (1.0 - NSE)
        return obj_value

def c1(metrics):
    NSE = float(metrics[0])
    return max(0.0, 0.5 - NSE)

def c2(metrics):
    TRMSE = float(metrics[1])
    return max(0.0, TRMSE - 1.6)

def c3(metrics):
    ROCE = float(metrics[2])
    return max(0.0, ROCE - 0.5)

def c4(metrics):
    SFDCE = float(metrics[3])
    return max(0.0, SFDCE - 0.2)
\end{formulationcode}

\formulationsubheading{Baselines}
The baselines include two expert-designed formulations and two direct LLM-generated formulations.
\formulationtitle[form:hbv-expert-a]{HBV $\mathcal{F}_{\text{Expert-A}}$}
\begin{formulationcode}
def obj1(metrics):
    try:
        import numpy as np
        arr = np.asarray(metrics, dtype=float)
        if arr.ndim == 1:
            return -float(arr[0]) + float(arr[2])
        if arr.ndim == 2 and arr.shape[0] >= 1:
            return -float(arr[0, 0]) + float(arr[0, 2])
        raise ValueError('metrics must be a 1D row or a 2D matrix with at least one row.')
    except Exception:
        row = [float(value) for value in metrics]
        return -float(row[0]) + float(row[2])

def c1(metrics):
    try:
        import numpy as np
        arr = np.asarray(metrics, dtype=float)
        if arr.ndim == 1:
            value = float(arr[0])
        elif arr.ndim == 2 and arr.shape[0] >= 1:
            value = float(arr[0, 0])
        else:
            raise ValueError('metrics must be a 1D row or a 2D matrix with at least one row.')
    except Exception:
        row = [float(value) for value in metrics]
        value = float(row[0])
    return 0.5 - value

def c2(metrics):
    try:
        import numpy as np
        arr = np.asarray(metrics, dtype=float)
        if arr.ndim == 1:
            value = float(arr[1])
        elif arr.ndim == 2 and arr.shape[0] >= 1:
            value = float(arr[0, 1])
        else:
            raise ValueError('metrics must be a 1D row or a 2D matrix with at least one row.')
    except Exception:
        row = [float(value) for value in metrics]
        value = float(row[1])
    return value - 1.6

def c3(metrics):
    try:
        import numpy as np
        arr = np.asarray(metrics, dtype=float)
        if arr.ndim == 1:
            value = float(arr[2])
        elif arr.ndim == 2 and arr.shape[0] >= 1:
            value = float(arr[0, 2])
        else:
            raise ValueError('metrics must be a 1D row or a 2D matrix with at least one row.')
    except Exception:
        row = [float(value) for value in metrics]
        value = float(row[2])
    return value - 0.5

def c4(metrics):
    try:
        import numpy as np
        arr = np.asarray(metrics, dtype=float)
        if arr.ndim == 1:
            value = float(arr[3])
        elif arr.ndim == 2 and arr.shape[0] >= 1:
            value = float(arr[0, 3])
        else:
            raise ValueError('metrics must be a 1D row or a 2D matrix with at least one row.')
    except Exception:
        row = [float(value) for value in metrics]
        value = float(row[3])
    return value - 0.2
\end{formulationcode}

\formulationtitle[form:hbv-expert-b]{HBV $\mathcal{F}_{\text{Expert-B}}$}
\begin{formulationcode}
def obj1(metrics):
    try:
        nse = float(metrics[0])
        roce = float(metrics[2])
    except Exception:
        nse = float(metrics[0][0])
        roce = float(metrics[0][2])
    if nse <= 0.0:
        return 1000000.0 + roce
    return -nse / (1.0 + roce)

def c1(metrics):
    try:
        value = float(metrics[0])
    except Exception:
        value = float(metrics[0][0])
    return 0.5 - value

def c2(metrics):
    try:
        value = float(metrics[1])
    except Exception:
        value = float(metrics[0][1])
    return value - 1.6

def c3(metrics):
    try:
        value = float(metrics[2])
    except Exception:
        value = float(metrics[0][2])
    return value - 0.5

def c4(metrics):
    try:
        value = float(metrics[3])
    except Exception:
        value = float(metrics[0][3])
    return value - 0.2
\end{formulationcode}

\formulationtitle[form:hbv-llm-a]{HBV $\mathcal{F}_{\text{LLM-A}}$}
\begin{formulationcode}
def obj1(metrics):
    NSE = metrics[0]
    TRMSE = metrics[1]
    ROCE = metrics[2]
    SFDCE = metrics[3]
    w_nse = 5.0
    w_roce = 3.0
    w_trmse = 1.0
    w_sfdce = 1.0
    t_nse = 0.5
    t_trmse = 1.6
    t_roce = 0.5
    t_sfdce = 0.2
    eps = 1e-08
    deficit_nse = max(0.0, t_nse - NSE)
    bonus_nse = 0.0
    if NSE >= t_nse:
        bonus_nse = -0.2 * (NSE - t_nse)
    cost_nse = deficit_nse + bonus_nse
    excess_trmse = max(0.0, TRMSE - t_trmse) / (t_trmse + eps)
    excess_roce = max(0.0, ROCE - t_roce) / (t_roce + eps)
    excess_sfdce = max(0.0, SFDCE - t_sfdce) / (t_sfdce + eps)
    base_trmse = TRMSE / (t_trmse + eps)
    base_roce = ROCE / (t_roce + eps)
    base_sfdce = SFDCE / (t_sfdce + eps)
    obj = w_nse * cost_nse + w_roce * (base_roce + excess_roce) + w_trmse * (base_trmse + 0.5 * excess_trmse) + w_sfdce * (base_sfdce + 0.5 * excess_sfdce)
    if deficit_nse == 0.0 and excess_trmse == 0.0 and (excess_roce == 0.0) and (excess_sfdce == 0.0):
        obj = obj - 2.0 * (NSE - t_nse) + 1.5 * (ROCE - t_roce)
    return float(obj)

def c1(metrics):
    NSE = metrics[0]
    TRMSE = metrics[1]
    ROCE = metrics[2]
    SFDCE = metrics[3]
    t_nse = 0.5
    t_trmse = 1.6
    t_roce = 0.5
    t_sfdce = 0.2
    v_nse = max(0.0, t_nse - NSE) / (abs(t_nse) + 1e-08)
    v_trmse = max(0.0, TRMSE - t_trmse) / (t_trmse + 1e-08)
    v_roce = max(0.0, ROCE - t_roce) / (t_roce + 1e-08)
    v_sfdce = max(0.0, SFDCE - t_sfdce) / (t_sfdce + 1e-08)
    max_violation = max(v_nse, v_trmse, v_roce, v_sfdce)
    return float(max_violation)
\end{formulationcode}

\formulationtitle[form:hbv-llm-b]{HBV $\mathcal{F}_{\text{LLM-B}}$}
\begin{formulationcode}
def obj1(metrics):
    nse = metrics[0]
    trmse = metrics[1]
    roce = metrics[2]
    sfdce = metrics[3]
    w_nse = 1.0
    w_roce = 1.0
    w_trmse = 0.3
    w_sfdce = 0.3
    target_nse_shift = 1.0
    val = w_nse * -(nse - -target_nse_shift) + w_roce * roce + w_trmse * trmse + w_sfdce * sfdce
    return float(val)

def c1(metrics):
    return float(0.5 - metrics[0])

def c2(metrics):
    return float(metrics[1] - 1.6)

def c3(metrics):
    return float(metrics[2] - 0.5)

def c4(metrics):
    return float(metrics[3] - 0.2)
\end{formulationcode}

\subsection{HSE Antenna Formulations}
\label{app:hse formulations}

For the antenna benchmark, we show the HSE task as the representative case. 
The formulation input is a sampled frequency-response curve represented as a two-column array, with frequency in the first column and response value in the second column.

\formulationsubheading{SHA-PF}
\formulationtitle[form:hse-shapf]{HSE SHA-PF}
\begin{formulationcode}
def obj1(curve):
    import numpy as np
    f = curve[:, 0]
    y = curve[:, 1]
    r1 = (f >= 0.8) & (f <= 0.92)
    r2 = (f >= 0.91) & (f <= 0.96)
    r3 = (f >= 0.95) & (f <= 1.08)
    r4 = (f >= 1.07) & (f <= 1.16)
    r5 = (f >= 1.15) & (f <= 1.2)

    def seg(mask):
        return y[mask]
    pb = seg(r3)
    if pb.size == 0:
        return 1000.0
    floor10 = np.quantile(pb, 0.1)
    floor25 = np.quantile(pb, 0.25)
    top75 = np.quantile(pb, 0.75)
    ripple = max(0.0, top75 - floor25)
    left_edge = y[(f >= 0.95) & (f < 0.97)]
    right_edge = y[(f > 1.06) & (f <= 1.08)]
    le = np.mean(left_edge) if left_edge.size > 0 else np.mean(pb)
    re = np.mean(right_edge) if right_edge.size > 0 else np.mean(pb)
    edge_def = max(0.0, -2.0 - min(le, re))
    curv = 0.0
    if pb.size >= 3:
        s2 = pb[:-2] - 2 * pb[1:-1] + pb[2:]
        curv = np.mean(np.maximum(np.abs(s2) - 0.2, 0.0))
    obj = -(0.7 * floor10 + 0.3 * floor25) + 0.6 * ripple + 0.4 * curv + 0.5 * edge_def

    def soft_over(x, thr, q=0.9, p=2.0):
        if x.size == 0:
            return 0.0
        t = np.quantile(x, q)
        v = max(0.0, t - thr)
        return v ** 1.0
    sb_pen = soft_over(seg(r1), -7.0, q=0.9) + soft_over(seg(r5), -7.0, q=0.9)

    def notch_shape(x):
        if x.size < 3:
            return 1.0
        i = int(np.argmin(x))
        m = x[i]
        vscore = 0.0
        if 1 <= i <= x.size - 2:
            sL = x[i] - x[i - 1]
            sR = x[i + 1] - x[i]
            vscore = max(0.0, sL) + max(0.0, -sR)
        else:
            vscore = 0.5
        left = x[:i] if i > 0 else np.array([])
        right = x[i + 1:] if i < x.size - 1 else np.array([])
        lc = 0.0 if left.size == 0 else max(0.0, m - np.max(left))
        rc = 0.0 if right.size == 0 else max(0.0, m - np.max(right))
        return 0.3 * vscore + 0.7 * (lc + rc)
    notch_guidance = 0.5 * notch_shape(seg(r2)) + 1.0 * notch_shape(seg(r4))
    return float(obj + 0.8 * sb_pen + 0.5 * notch_guidance)

def c1(curve):
    import numpy as np
    f = curve[:, 0]
    y = curve[:, 1]
    r1 = (f >= 0.8) & (f <= 0.92)
    r3 = (f >= 0.95) & (f <= 1.08)
    r4 = (f >= 1.07) & (f <= 1.16)
    r5 = (f >= 1.15) & (f <= 1.2)

    def q(x, qv, defv):
        return np.quantile(x, qv) if x.size > 0 else defv
    v_sbL = 5 * (q(y[r1], 0.95, -7.0) - -7.0)
    v_sbH = q(y[r5], 0.95, -7.0) - -7.0
    v_pb = -2.0 - q(y[r3], 0.05, -2.0)
    segH = y[r4]

    def notch_violation(seg, thr):
        if seg.size < 3:
            return 1.0
        m = np.min(seg)
        depth_v = m - thr
        dy = np.diff(seg)
        eps = 0.0001
        has_min = np.any((dy[:-1] < -eps) & (dy[1:] > eps))
        mono_v = 0.5 if not has_min else 0.0
        return max(depth_v, mono_v)
    v_nH = notch_violation(segH, -14.0)
    return float(max(v_sbL, v_sbH, v_pb, v_nH))

def c2(curve):
    import numpy as np
    f = curve[:, 0]
    y = curve[:, 1]
    r2 = (f >= 0.91) & (f <= 0.96)
    seg = y[r2]
    if seg.size < 3:
        return 1.0
    m = np.min(seg)
    dy = np.diff(seg)
    eps = 0.0001
    has_min = np.any((dy[:-1] < -eps) & (dy[1:] > eps))
    mono_v = 7 if not has_min else 0.0
    depth_v = m - -8.0 if m > -8.0 else 0.0
    return float(max(depth_v, mono_v))
\end{formulationcode}

\formulationsubheading{Baselines}
The baselines again include two expert-designed formulations and two direct LLM-generated formulations.
\formulationtitle[form:hse-expert-a]{HSE $\mathcal{F}_{\text{Expert-A}}$}
\begin{formulationcode}
def obj1(curve):
    import numpy as np
    f = curve[:, 0]
    y = curve[:, 1]
    mask = (f >= 0.95) & (f <= 1.07)
    if np.any(mask):
        return float(-np.min(y[mask]))
    return float(0.0)

def c1(curve):
    import numpy as np
    f = curve[:, 0]
    y = curve[:, 1]
    mask = (f >= 0.91) & (f <= 0.96)
    if np.any(mask):
        return float(np.min(y[mask]) - -8.0)
    return float(1000.0)

def c2(curve):
    import numpy as np
    f = curve[:, 0]
    y = curve[:, 1]
    mask = (f >= 1.07) & (f <= 1.16)
    if np.any(mask):
        return float(np.min(y[mask]) - -14.0)
    return float(1000.0)

def c3(curve):
    import numpy as np
    f = curve[:, 0]
    y = curve[:, 1]
    mask = (f >= 0.8) & (f <= 0.92)
    if np.any(mask):
        return float(np.max(y[mask]) - -7.0)
    return float(1000.0)

def c4(curve):
    import numpy as np
    f = curve[:, 0]
    y = curve[:, 1]
    mask = (f >= 1.15) & (f <= 1.2)
    if np.any(mask):
        return float(np.max(y[mask]) - -7.0)
    return float(1000.0)
\end{formulationcode}

\formulationtitle[form:hse-expert-b]{HSE $\mathcal{F}_{\text{Expert-B}}$}
\begin{formulationcode}
def obj1(curve):
    import numpy as np
    f = curve[:, 0]
    y = curve[:, 1]
    pass_mask = (f >= 0.95) & (f <= 1.07)
    low_stop_mask = (f >= 0.8) & (f <= 0.92)
    high_stop_mask = (f >= 1.15) & (f <= 1.2)
    if not np.any(pass_mask):
        return float(0.0)
    value = -np.min(y[pass_mask])
    if np.any(low_stop_mask):
        value += 0.5 * np.max(y[low_stop_mask])
    if np.any(high_stop_mask):
        value += 0.5 * np.max(y[high_stop_mask])
    return float(value)

def c1(curve):
    import numpy as np
    f = curve[:, 0]
    y = curve[:, 1]
    mask = (f >= 0.95) & (f <= 1.07)
    if np.any(mask):
        return float(-2.0 - np.min(y[mask]))
    return float(1000.0)

def c2(curve):
    import numpy as np
    f = curve[:, 0]
    y = curve[:, 1]
    mask = (f >= 0.91) & (f <= 0.96)
    if np.any(mask):
        return float(np.min(y[mask]) - -8.0)
    return float(1000.0)

def c3(curve):
    import numpy as np
    f = curve[:, 0]
    y = curve[:, 1]
    mask = (f >= 1.07) & (f <= 1.16)
    if np.any(mask):
        return float(np.min(y[mask]) - -14.0)
    return float(1000.0)
\end{formulationcode}

\formulationtitle[form:hse-llm-a]{HSE $\mathcal{F}_{\text{LLM-A}}$}
\begin{formulationcode}
def obj1(curve):
    import numpy as np
    f = curve[:, 0]
    y = curve[:, 1]
    idx_pb = (f >= 0.95) & (f <= 1.07)
    if not np.any(idx_pb):
        return 1000000.0
    mean_pb = np.mean(y[idx_pb])
    idx_ls = (f >= 0.8) & (f <= 0.92)
    idx_hs = (f >= 1.15) & (f <= 1.2)
    reg_ls = np.mean(np.maximum(y[idx_ls], -100)) if np.any(idx_ls) else 0.0
    reg_hs = np.mean(np.maximum(y[idx_hs], -100)) if np.any(idx_hs) else 0.0
    return -mean_pb + 0.02 * (reg_ls + reg_hs)

def c1(curve):
    import numpy as np
    f = curve[:, 0]
    y = curve[:, 1]
    idx = (f >= 0.8) & (f <= 0.92)
    if not np.any(idx):
        return 0.0
    return np.maximum(0.0, np.max(y[idx]) - -7)

def c2(curve):
    import numpy as np
    f = curve[:, 0]
    y = curve[:, 1]
    idx = (f >= 0.91) & (f <= 0.96)
    if not np.any(idx) or len(y[idx]) < 3:
        return 5.0
    ys = y[idx]
    fs = f[idx]
    local_min_exists = False
    for i in range(1, len(ys) - 1):
        if ys[i] <= ys[i - 1] - 0.5 and ys[i] <= ys[i + 1] - 0.5:
            local_min_exists = True
            min_val = ys[i]
            break
    if not local_min_exists:
        diffs = np.diff(ys)
        score = max(np.max(np.maximum(diffs, 0.0)), -np.min(np.minimum(diffs, 0.0))) if len(diffs) > 0 else 5.0
        return 1.0 + score
    return np.maximum(0.0, min_val - -8)

def c3(curve):
    import numpy as np
    f = curve[:, 0]
    y = curve[:, 1]
    idx = (f >= 0.95) & (f <= 1.07)
    if not np.any(idx):
        return 5.0
    min_pb = np.min(y[idx])
    return np.maximum(0.0, -2 - min_pb)

def c4(curve):
    import numpy as np
    f = curve[:, 0]
    y = curve[:, 1]
    idx = (f >= 1.07) & (f <= 1.16)
    if not np.any(idx) or len(y[idx]) < 3:
        return 3.0
    ys = y[idx]
    fs = f[idx]
    local_min_exists = False
    for i in range(1, len(ys) - 1):
        if ys[i] <= ys[i - 1] - 0.3 and ys[i] <= ys[i + 1] - 0.3:
            local_min_exists = True
            min_val = ys[i]
            break
    if not local_min_exists:
        diffs = np.diff(ys)
        score = max(np.max(np.maximum(diffs, 0.0)), -np.min(np.minimum(diffs, 0.0))) if len(diffs) > 0 else 3.0
        return 0.5 + score
    return np.maximum(0.0, min_val - -14)

def c5(curve):
    import numpy as np
    f = curve[:, 0]
    y = curve[:, 1]
    idx = (f >= 1.15) & (f <= 1.2)
    if not np.any(idx):
        return 0.0
    return np.maximum(0.0, np.max(y[idx]) - -7)
\end{formulationcode}

\formulationtitle[form:hse-llm-b]{HSE $\mathcal{F}_{\text{LLM-B}}$}
\begin{formulationcode}
def obj1(curve):
    import numpy as np
    f = curve[:, 0]
    y = curve[:, 1]
    idx_pb = (f >= 0.95) & (f <= 1.07)
    if not np.any(idx_pb):
        return 1000000.0
    mean_pb = np.mean(y[idx_pb])
    idx_n1 = (f >= 0.91) & (f <= 0.96)
    idx_n2 = (f >= 1.07) & (f <= 1.16)
    min_n1 = np.min(y[idx_n1]) if np.any(idx_n1) else 0.0
    min_n2 = np.min(y[idx_n2]) if np.any(idx_n2) else 0.0
    idx_s1 = (f >= 0.8) & (f <= 0.92)
    idx_s2 = (f >= 1.15) & (f <= 1.2)
    mean_s1 = np.mean(y[idx_s1]) if np.any(idx_s1) else -100.0
    mean_s2 = np.mean(y[idx_s2]) if np.any(idx_s2) else -100.0
    return -mean_pb + 0.05 * (min_n1 - -8 + (min_n2 - -14)) + 0.01 * (mean_s1 + mean_s2)

def c1(curve):
    import numpy as np
    f,y = curve[:, 0], curve[:, 1]
    ys = y
    fs = f
    if len(ys) < 5:
        return 0.0
    d1 = np.diff(ys) / np.diff(fs)
    d2 = np.diff(d1) / (fs[2:] - fs[1:-1])
    max_curv = np.max(np.abs(d2))
    return np.maximum(0.0, max_curv - 50.0)

def c2(curve):
    import numpy as np
    f = curve[:, 0]
    y = curve[:, 1]
    idx_pb = (f >= 0.95) & (f <= 1.07)
    if not np.any(idx_pb):
        return 5.0
    min_pb = np.min(y[idx_pb])
    return np.maximum(0.0, 0.0 - min_pb)

def c3(curve):
    import numpy as np
    f = curve[:, 0]
    y = curve[:, 1]

    def has_notch(idx_band, thresh_depth):
        ys = y[idx_band]
        fs = f[idx_band]
        if len(ys) < 3:
            return (False, 10.0)
        for i in range(1, len(ys) - 1):
            if ys[i] <= ys[i - 1] - 0.3 and ys[i] <= ys[i + 1] - 0.3:
                return (True, ys[i])
        return (False, np.min(ys))
    idx1 = (f >= 0.91) & (f <= 0.96)
    idx2 = (f >= 1.07) & (f <= 1.16)
    (ok1, min1) = has_notch(idx1, -8)
    (ok2, min2) = has_notch(idx2, -14)
    pen1 = 0.0 if ok1 else 2.0
    pen1 += np.maximum(0.0, min1 - -8)
    pen2 = 0.0 if ok2 else 2.0
    pen2 += np.maximum(0.0, min2 - -14)
    return pen1 + pen2
\end{formulationcode}

\end{document}